\documentclass[10pt,journal,compsoc]{IEEEtran}

\usepackage{graphicx, epstopdf, amsmath, float, amssymb, amsthm, mathtools}
\usepackage{subfig}
\usepackage[ruled]{algorithm2e}
\def\R{\mathbb{R}}

\usepackage{url}
\usepackage[usenames,dvipsnames]{color}
\usepackage[colorlinks=true]{hyperref}
\hypersetup{
  colorlinks,
 citecolor=Blue,
 linkcolor=Red,
 urlcolor=Violet}

\ifCLASSOPTIONcompsoc
  \usepackage[nocompress]{cite}
\else
  \usepackage{cite}
\fi
\usepackage{array}
\begin{document}
\title{von Mises-Fisher Mixture Model-based Deep learning: Application to Face Verification}

\author{Md.~Abul~Hasnat,
		Julien~Bohn\'{e},
		Jonathan~Milgram,	
		St\'{e}phane~Gentric
        and~Liming~Chen
\thanks{Laboratoire LIRIS, \'{E}cole centrale de Lyon, 69134 Ecully, France.}%
\thanks{Safran Identity \& Security, 92130 Issy-les-Moulineaux, France.}
\thanks{e-mail:md-abul.hasnat@ec-lyon.fr, julien.bohne@safrangroup.com,}%
\thanks{jonathan.milgram@safrangroup.com, stephane.gentric@safrangroup.com}%
\thanks{liming.chen@ec-lyon.fr}}

\maketitle

\begin{abstract}
A number of pattern recognition tasks, \textit{e.g.}, face verification, can be boiled down to classification or clustering of unit length directional feature vectors whose distance can be simply computed by their angle. In this paper, we propose the von Mises-Fisher (vMF) mixture model as the theoretical foundation for an effective deep-learning of such directional features and derive a novel vMF Mixture Loss and its corresponding vMF deep features. The proposed vMF feature learning achieves the characteristics of discriminative learning, \textit{i.e.}, compacting the instances of the same class while increasing the distance of instances from different classes. Moreover, it subsumes a number of popular loss functions as well as an effective method in deep learning, namely normalization. We conduct extensive experiments on face verification using 4 different challenging face datasets, \textit{i.e.}, LFW, YouTube faces, CACD and IJB-A. Results show the effectiveness and excellent generalization ability of the proposed approach as it achieves state-of-the-art results on the LFW, YouTube faces and CACD datasets and competitive results on the IJB-A dataset.
\end{abstract}


\IEEEpeerreviewmaketitle
%
%
\section{Introduction}
\label{sec:introduction}
A number of pattern recognition tasks, \textit{e.g.}, face recognition (FR), can be boiled down to supervised classification or unsupervised clustering of unit length feature vectors whose distance can be simply computed by their angle, \textit{i.e.}, cosine distance. In deep learning based FR, numerous methods find it useful to unit-normalize the final feature vectors, \textit{e.g.}, \cite{wang2016face, ding2015robust, centerlosswen2016, yi2014learning, spherefaceliu2017, normfacewang2017, l2ranjan2017}. Besides, the widely used and simple softmax loss has been extended with additional or reinforced supervising signal, \textit{e.g.}, center loss \cite{centerlosswen2016}, large margin softmax loss \cite{liu2016large}, to further enable a discriminative learning, \textit{i.e.}, compacting intra-class instances while repulsing inter-class instances, and thereby increase the final recognition accuracy. 
However, the great success of these methods and practices remains unclear from a theoretical viewpoint, which motivates us to \textit{study the deep feature representation from a theoretical perspective}.

Statistical Mixture Models (MM) is a common method to perform probabilistic clustering and widely used in data mining and machine learning \cite{murphy2012machine}. MM plays key role in model based clustering \cite{biernacki2000assessing}, which assumes a generative model, \textit{i.e.}, each observation is a sample from a finite mixture of probability distributions. 
In this paper, we adopt the \textit{theoretical concept of MM to model the deep feature representation task} and realize the relationship among the model parameters and deep features.

\begin{figure}[!t]
\centering
\subfloat[]{\includegraphics[scale=0.29]{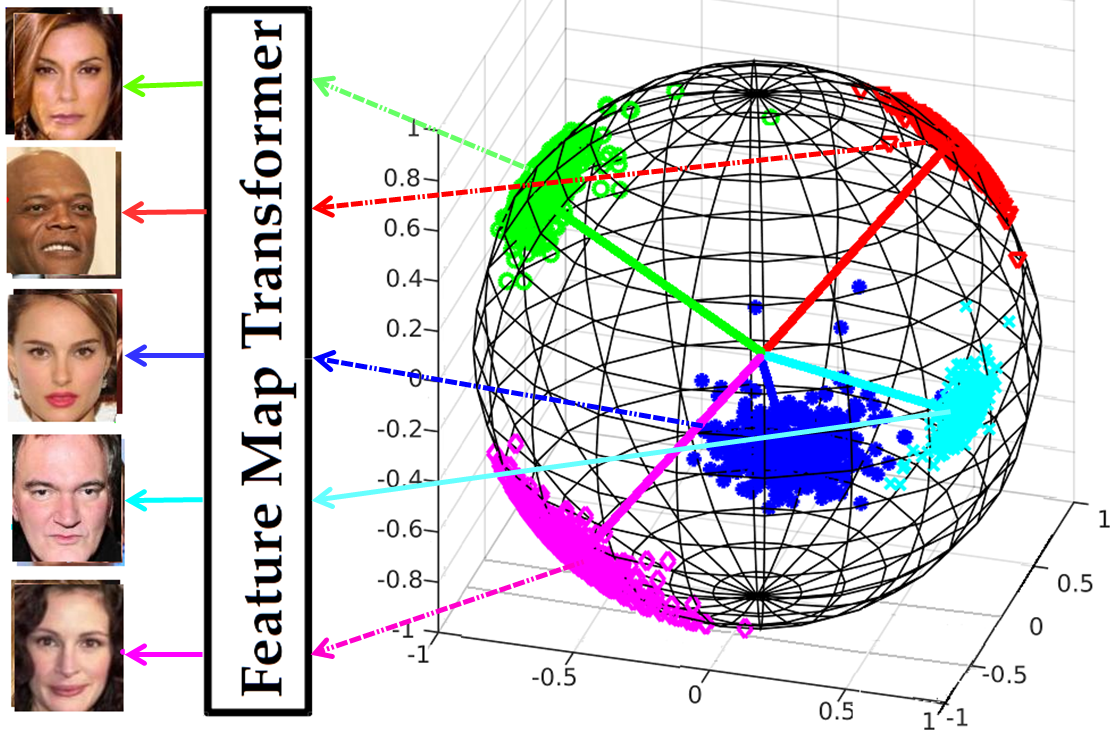}}
\hfill
\subfloat[]{\includegraphics[scale=0.37]{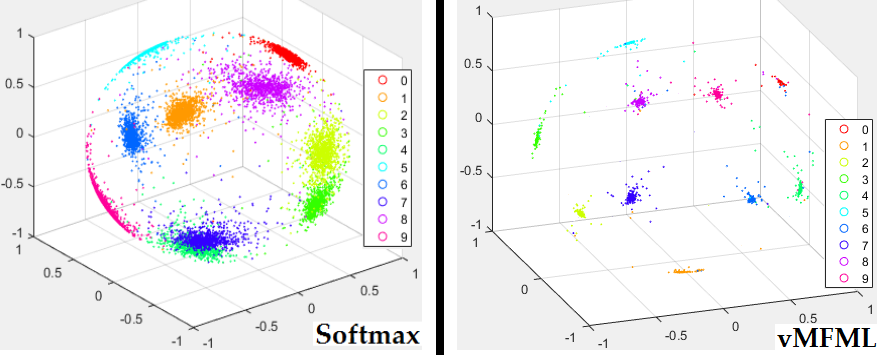}}
\caption{(a) illustration of the proposed model with a 5 classes vMFMM, where \textit{facial identity} represents the \textit{classes} and (b) illustration of the 3D features learned from the MNIST digits \cite{lecun1998gradient}, \textit{left}: softmax loss, \textit{right}: proposed (vMFML). Features from different classes are shown with different markers and colors in the sphere. See Appendix \ref{secapp:proof_concept} for an extended experiment and further illustrations.}
\label{fig:ffr_model}
\end{figure}

Unit length normalized feature vectors are directional features which only keep the orientations of data features as discriminative information while ignoring their magnitude. In this case, simple angle measurement, \textit{e.g.}, cosine distance, can be used as dissimilarity measure of two data points and provides very intuitive geometric interpretation of similarity \cite{gopal2014mises}. In this paper, we propose to model the (deep)-features delivered by the deep neural nets, \textit{e.g.}, CNN-based neural networks, as a mixture of von Mises-Fisher distributions, also called vMF Mixture Model (vMFMM). 
The von Mises-Fisher (vMF) is a fundamental probability distribution, which has been successfully used in numerous unsupervised classification tasks \cite{hasnat2016joint, banerjee2005clustering, gopal2014mises}. By combining this vMFMM with deep neural networks, we derive a novel loss function, namely vMF mixture loss (vMFML) which enables a discriminative learning. Figure \ref{fig:ffr_model}(a) (from right to left) provides an illustration of the proposed model. Figure \ref{fig:ffr_model}(b) shows\footnote{We visualize the features of the MNIST digits in the 3D space (similar illustration with 2D plot in \cite{centerlosswen2016}). The CNN is composed of 6 convolution, 2 pool and 1 FC (with 2 neurons for 2D visualization) layers. We optimize it using the softmax loss and the proposed vMFML.} the discriminative nature of the proposed model, \textit{i.e.}, the learned features for each class are compacted whereas inter-class features are repulsed.

To demonstrate the effectiveness of the proposed method, we carried out extensive experiments on face recognition (FR) task on which recent deep learning-based methods \cite{taigman2014deepface, schroff2015facenet, sun2015deepid3, parkhi2015deep, baiduliu2015} have surpassed human level performance. We used 4 different challenging face datasets, namely LFW \cite{lfw_huang2014} for single image based face verification, IJB-A \cite{ijbaKlare2015} for face templates matching,  YouTube Faces \cite{ytfwolf2011} (YTF)  for video faces matching, and CACD \cite{chen2015facecacd}  for cross age face matching. Using only one deep CNN model trained on the MS-Celeb dataset \cite{mscelebguo16}, the proposed method achieves 99.63\% accuracy on LFW, 85\%  TAR@FAR=0.001 on IJB-A \cite{ijbaKlare2015}, 96.46\% accuracy on YTF \cite{ytfwolf2011} and 99.2\% accuracy on CACD \cite{chen2015facecacd}. These results indicate that our method generalizes very well across different datasets as it achieves state-of-the-art results on the LFW, YouTube faces and CACD datasets and competitive results on the IJB-A dataset.

The contributions of the proposed method can be summarized as follows:
\begin{itemize}
\item we propose a novel \textit{feature representation} model (Sect. \ref{sssec:ffrm}) from a theoretical perspective. It comprises the statistical mixture model \cite{murphy2012machine} with directional distribution \cite{mardia2009directional}. It provides a novel view to understand and model the desired pattern classification task. Therefore, it can help to develop efficient methods to achieve better results. 
\item we propose a \textit{directional feature representation learning} method, called vMF-FL (Sect. \ref{sssec:frlm}), which combines the theoretical model with the CNN model. vMF-FL provides a novel loss function, called vMFML (Sect. \ref{ssec:ffr_vmf_loss}), whose formulation \textit{w.r.t.} the backpropagation \cite{lecun1998gradient} method shows that it can be easily integrated with any CNN model. Moreover, vMFML is able to explain (Sect. \ref{int_discussion}) different loss functions \cite{centerlosswen2016, liu2016large, spherefaceliu2017, normfacewang2017, l2ranjan2017} and normalization methods \cite{weight_normalization, cosine_normalization, batch_normalization}. vMFML not only interprets the relation among the parameters and features, but also improves the CNN learning task \textit{w.r.t.} efficiency (faster convergence) and performance (better accuracy, Sect. \ref{sss:comp_loss_func}). Therefore, it can be used in a variety of classification tasks under the assumption of directional features.
\item we verify (Sect. \ref{ssec:res_eval}) the effectiveness of the proposed vMFML on the task of face verification through comprehensive experiments on four face benchmarks depicting various challenges, \textit{e.g.}, pose, lighting, age, and demonstrate its generalization skills across datasets. 

\item we perform (sect.\ref{ssec:disscussion}) additional experiments and provide in-depth analysis of various factors, \textit{e.g.}, sensitivity of parameters, training data size, failure cases,  to further gain insight of the proposed method.
\end{itemize} 

In the remaining part, we study the related work in Section \ref{sec:rel_work}, describe our method in Section \ref{sec:methodology}, present experimental results with analysis in Section \ref{sec:res_exp} and finally draw conclusions in Section \ref{sec:conclusion}.

\section{Related Work}
\label{sec:rel_work}
\subsection{Mixture models and Loss Functions}
\label{ssec:rel_work_mm_loss}
\textit{\textbf{Mixture models (MM)}} \cite{murphy2012machine} have been widely used in numerous machine learning problems, such as classification \cite{fernando2012supervised}, clustering \cite{biernacki2000assessing}, image analysis \cite{hasnat2016joint}, text analysis \cite{banerjee2005clustering}, shape retrieval \cite{liu2012shape}, \textit{etc}. However, their potentials are relatively under-explored with the neural network (NN) based learning methods. Several recent work \cite{van2014factoring, patel2016probabilistic, variani2015gaussian, tuske2015integrating} explored the concept of MM with NN from different perspectives. \cite{van2014factoring} used the Gaussian MM (GMM) to model deep NN as a mixture of transformers. \cite{patel2016probabilistic} aimed to capture the variations in the nuisance variables (\textit{e.g.}, object poses) and used NN as a rendering function to propose deep rendering MM. Both of these methods use MM within the NN,
whereas \textit{we consider NN as a single transformer}. \cite{variani2015gaussian, tuske2015integrating} used GMM with NN and applied it for speech analysis. \cite{variani2015gaussian} performed discriminative feature learning task via the proposed GMM layer. \cite{tuske2015integrating} used the concept of log-linear model with GMM and modified the softmax loss accordingly. While our method is more similar to \cite{variani2015gaussian, tuske2015integrating}, there are several differences: (a) we use directional (unit normalized) features; (b) we use the vMF \cite{mardia2009directional} distribution which is more appropriate for directional features \cite{hasnat2016model}; (c) our feature representation model is based on a generative model-based concept; and (d) we exploit the CNN model and explore practical application of computer vision. 

\textit{\textbf{MM with directional distributions}} \cite{mardia2009directional} have been used in a variety of domains to analyze images \cite{hasnat2016model}, speech \cite{vu2010blind}, text \cite{banerjee2005clustering, gopal2014mises}, gene expressions \cite{banerjee2005clustering}, shapes \cite{prati2008using}, pose \cite{glover2012monte}, diffusion MRI \cite{bhalerao2007hyperspherical}, \textit{etc}. However, they remain unexplored to learn discriminative features. In this paper, we aim to explore it by modeling the feature learning task with the vMF distribution \cite{mardia2009directional, banerjee2005clustering} and combining it with a CNN model. To the best of our knowledge, \textit{this is the first reported attempt to use the directional distribution with the CNN model}.

\textit{\textbf{Loss functions}} are essential part of CNN training. 
The CNN model parameters are learned by optimizing certain loss functions which are defined based on the given task (\textit{e.g.}, classification, regression) and the available supervisory information (\textit{e.g.}, class labels, price). 
While the softmax loss \cite{recentcnngu2015} is most commonly incorporated, recent researches \cite{centerlosswen2016, liu2016large} indicate that it cannot guarantee to provide discriminative features. \cite{centerlosswen2016} proposed center loss as a supplementary loss to minimize intra-class variations and hence to improve feature discrimination. 
Our method \textit{performs the same task with the concentration ($\kappa$) parameter and requires no supplementary loss function}, see Section \ref{int_discussion} for details.
\cite{liu2016large, spherefaceliu2017} proposed the large-margin softmax loss by incorporating an intuitive margin on the classification boundary, which can be explained by our method under certain condition, see Section \ref{int_discussion} for details.
\subsection{Face Recognition}
\label{ssec:rel_work_fr}
\begin{table*}[!t]
\centering
\caption{Overview of the state-of-the-art FR methods. In the second column (\textbf{CNN Info}), the shorthand notations mean- \textbf{C}- convolutional layer, \textbf{FC}: fully connected layer, \textbf{LC}: locally connected layer and \textbf{L}: loss layer. In the third column (\textbf{Loss Function}), the shorthand notations mean- \textbf{TL}: triplet loss, \textbf{SL}: softmax loss, \textbf{CL}: contrastive loss,  \textbf{CCL}: C-contrastive loss, \textbf{CeL}: center loss, \textbf{ASL}: angular softmax loss and \textbf{RL}: range loss. In the fourth column (\textbf{Additional Learning}), the shorthand notations mean- \textbf{JB}: joint bayesian, \textbf{TSE}: triplet similarity embedding,  \textbf{TPE}: triplet probability embedding, \textbf{TA}: template adaptation and \textbf{X}: no metric learning. We list the methods in a decreasing order based on the number of convolution and FC layers in the CNN model.}
\label{tab:cnn_analysis}
\begin{tabular}{|c|c|c|c|c|c|}
\hline
\textbf{FR system}                         & \textbf{CNN Info}                                                              & \textbf{\begin{tabular}[c]{@{}c@{}}Loss\\ function\end{tabular}} & \textbf{\begin{tabular}[c]{@{}c@{}}Additional\\ learning\end{tabular}} & \textbf{\begin{tabular}[c]{@{}c@{}}\# of\\ CNNs\end{tabular}} & \textbf{\begin{tabular}[c]{@{}c@{}}Dataset\\ Info\end{tabular}} \\ \hline
DeepID2+ \cite{deepid2psun2015}  & 4-C, 1-FC, 2L                                                                                                                   & SL, CL & JB & 25  & 0.29M, 12K
\\ \hline
Deepface \cite{taigman2014deepface}      & 2-C, 3-LC, 1-FC, 1-L                                                                & SL, CL & X & 3 & 4.4M, 4K
\\ \hline
Webscale \cite{webscaletaigman2015} & 2-C, 3-LC, 2-FC, 1-L & SL & X & 4 & 4.5M, 55K
\\ \hline
Center Loss \cite{centerlosswen2016} & 3-C, 3-LC, 1-FC, 2-L & SL, CeL & X & 1 & 0.7M, 17.2K    
\\ \hline
FV-TSE \cite{Sankaranarayanan2016} & 6-C, 2-FC, 1-L & SL & TSE & 1 & 0.49M, 10.5K
\\ \hline
FV-TPE \cite{Sankaranarayanan2016a} & 7-C, 2-FC, 1-L & SL & TPE & 1 & 0.49M, 10.5K
\\ \hline
VIPLFaceNet	\cite{viplfacenetliu2016} & 7-C, 2-FC, 1-L & SL & X & 1 & 0.49M, 10.57K
\\ \hline
All-In-One \cite{allinoneranjan2016} & 7-C, 2-FC, 8-L & SL & TPE & 1 & 0.49M, 10.5K
\\ \hline
CASIA-Webface \cite{yi2014learning}     & 10-C, 1-L & SL & JB & 1 & 0.49M, 10.5K
\\ \hline
FSS \cite{wang2016face} & 10-C, 1-L & SL & JB & 9 & 0.49M, 10.57K
\\ \hline
Unconstrained FV \cite{chen2016unconstrained} & 10-C, 1-FC, 1-L & SL & JB & 1 & 0.49M, 10.5K
\\ \hline
Sparse ConvNet \cite{sparsifyingsun2015} & 10-C, 1-FC, 1-L & SL, CL & JB & 25 & 0.29M, 12K
\\ \hline
FaceNet \cite{schroff2015facenet}        & 11-C, 3-FC, 1-L & TL                                                            & X & 1 & 200M, 8M
\\ \hline
DeepID3 \cite{sun2015deepid3}            & 8-C, 4-FC, 2-LC, 2L                                                                                                                   & SL, CL & JB & 25  & 0.3M, 13.1K \\ \hline
MM-DFR \cite{ding2015robust}             & \begin{tabular}[c]{@{}c@{}}10-C, 2-FC, 1-L and \\ 12-C, 2-FC, 1-L\end{tabular} & SL & JB & 8 & 0.49M, 10.5K \\ \hline
VGG Face \cite{parkhi2015deep}           & 13-C, 3-FC, 2L                                                                 & SL, TL & X & 1 & 2.6M, 2.6K \\ \hline
FV-TA \cite{Crosswhite2016} & 13-C, 3-FC, 2L & SL, TL & TA & 1 & 2.6M, 2.6K 
\\ \hline
MFM-CNN \cite{wu2015lightened}           & 14-C, 2-FC, 1L                                                                                                                   & SL & X & 1 & 0.49M, 10.5K
\\ \hline
Face-Aug-Pose-Syn \cite{masi16dowe} & 16-C, 3-FC, 1-L & SL & PCA & 1 & 0.49M, 10.57K 
\\ \hline
Deep Multipose \cite{abdalmageed2016face} & \begin{tabular}[c]{@{}c@{}}16-C, 3-FC, 1-L and \\ 5-C, 2-FC, 1-L\end{tabular}  & SL & PCA & 12 & 0.4M, 10.5K
\\ \hline
Pose aware FR \cite{masi2016pose} & 16-C, 3-FC, 1-L & SL & PCA & 5 & 0.49M, 10.5K
\\ \hline
Range Loss \cite{rangelosszhang2016} & 27-C, 1-FC, 2-L & SL, RL & X & 1 & 1.5M, 100K    
\\ \hline
DeepVisage \cite{deepvisage_2017_ICCVW} & 27-C, 1-FC, 1-L & SL & X & 1 & 4.48M, 62K    
\\ \hline
NormFace \cite{normfacewang2017} & 27-C, 1-FC, 2-L & SL, CCL & X & 1 & 0.49M, 10.5K
\\ \hline
Megvii \cite{zhou2015naive}              & 4 $\times$ 10, 1-L & SL                                                              & X & 1 & 5M, 0.2M 
\\ \hline
Baidu \cite{baiduliu2015}        & 4 $\times$ 9-C, 2-L & SL, TL                                                            & X & 10 & 1.2M, 1.8K
\\ \hline
NAN \cite{nanyang2016} & 57-C, 5-FC, 3-L & SL, CL & Agg & 1 & 3M, 50K
\\ \hline
SphereFace \cite{spherefaceliu2017} & 64-C, 1-FC, 1-L & ASL & X & 1 & 0.49M, 10.5K
\\ \hline
L2-Softmax \cite{l2ranjan2017} & 100-C, 1-FC, 1-L & L2-SL & X & 1 & 0.49M, 10.5K
\\ \hline
\end{tabular}
\end{table*}
\textit{\textbf{Face recognition}} (FR) is a widely studied problem in which remarkable results have been achieved by the recent deep CNN based methods on several standard FR benchmarks datasets, such as, the Labeled Faces in the Wild (LFW) \cite{lfw_huang2014}, YouTube Faces (YTF) \cite{ytfwolf2011} and IARPA Janus Benchmark A (IJB-A) \cite{ijbaKlare2015}. In Table \ref{tab:cnn_analysis}, we study\footnote{We only consider the deep CNN based methods. For the others, we refer readers to the survey \cite{learned2016labeled} for LFW and the \cite{ijbaKlare2015} for IJB-A.} these methods and decompose their key aspects as: (a) CNN model design; (b) objective/loss functions; (c) fine-tune and additional learning method; (d) multi-modal input and number of CNNs and (e) use of the training database.

Recent deep CNN based FR methods tend to adopt (directly or slightly modify) the famous CNN architectures \cite{recentcnngu2015} proposed for the ImageNet \cite{imagenet_ijcv} challenge. The \emph{CNN Info} column of Table \ref{tab:cnn_analysis} provides the details of CNN models used by different FR methods. The AlexNet model is used by \cite{Sankaranarayanan2016, Sankaranarayanan2016a, viplfacenetliu2016, allinoneranjan2016, abdalmageed2016face, masi2016pose, schroff2015facenet}, VGGNet model is used by \cite{parkhi2015deep, Crosswhite2016, masi16dowe, abdalmageed2016face, masi2016pose, ding2015robust, sparsifyingsun2015}, GoogleNet model is used by \cite{nanyang2016, schroff2015facenet} and ResNet model is used by \cite{rangelosszhang2016, centerlosswen2016, normfacewang2017,spherefaceliu2017,l2ranjan2017,deepvisage_2017_ICCVW}. Besides the famous CNN models, \cite{yi2014learning} proposed a simpler CNN model which is used by \cite{wang2016face, chen2016unconstrained, ding2015robust}. 
CNNs with lower depth have been used by \cite{deepid2psun2015, taigman2014deepface, webscaletaigman2015, centerlosswen2016, sun2015deepid3}, where the model complexity is increased with locally connected convolutional layers. Parallel CNNs have been employed by \cite{baiduliu2015, zhou2015naive} to simultaneously learn features from different facial regions.
\textit{We adopt the ResNet \cite{resnethe2015deep} based deeper CNN model}.

FR methods are trained with different loss functions, see \emph{Loss function} column in Table \ref{tab:cnn_analysis}. Most FR methods learn the facial feature representation model by training the CNN for identity classification. For this purpose, the softmax loss \cite{recentcnngu2015} is used to optimize the classification objective, which requires the facial images with associated identity labels. 
Several variants of the softmax loss \cite{l2ranjan2017, liu2016large, spherefaceliu2017,normfacewang2017} have been recently proposed to enhance its discriminability. Besides, the contrastive loss \cite{recentcnngu2015, contrastive_loss} is used by \cite{deepid2psun2015, taigman2014deepface, sparsifyingsun2015, sun2015deepid3, nanyang2016} for face verification, which requires similar/dissimilar face image pairs and similarity labels. Moreover, the triplet loss \cite{schroff2015facenet} is used by \cite{schroff2015facenet, parkhi2015deep, Crosswhite2016, baiduliu2015} for face verification and requires the face triplets (\textit{i.e.} anchor, positive, negative). 
Recently, \cite{normfacewang2017} proposed different formulations of the contrastive and triplet losses such that they can be trained with only the identity labels. \textit{Our proposed vMFML, simply learns the features via identity classification and requires only the class labels}. Moreover, the theoretical foundation of vMFML provides interesting interpretation and relationship with the very recently proposed softmax based loss functions, such as \cite{l2ranjan2017, liu2016large, spherefaceliu2017, normfacewang2017}, see Sect. \ref{int_discussion} for details.

CNN training with multiple loss functions have been adopted by several methods, where the losses are optimized jointly \cite{deepid2psun2015, sparsifyingsun2015, sun2015deepid3, centerlosswen2016, rangelosszhang2016, allinoneranjan2016} or sequentially (optimize softmax loss followed by the other loss) \cite{taigman2014deepface, parkhi2015deep, Crosswhite2016, baiduliu2015, nanyang2016}. However, multiple loss optimization not only requires additional efforts for training data preparation but also complicates the CNN training procedure \cite{deepvisage_2017_ICCVW}. \textit{Our FR method optimizes single loss and do not need extra efforts}.

After training the CNN model, a second learning method is often incorporated by different FR methods, see \emph{Additional Learning} column of Table \ref{tab:cnn_analysis}. CNN fine-tune is a particular form of transfer learning commonly employed by several methods \cite{wang2016face, chen2016unconstrained, Sankaranarayanan2016}. It consists of updating the trained CNN parameters using the target specific training data. In order to adapt the CNN features to particular FR tasks, several methods apply additional transformation on the CNN features based on the \textit{metric/distance learning} strategy. The Joint Bayesian method \cite{chen2012bayesian} is a popular metric learning method and used by \cite{deepid2psun2015, yi2014learning, wang2016face, chen2016unconstrained, sparsifyingsun2015, sun2015deepid3, ding2015robust}. Specific embedding learning methods \cite{Sankaranarayanan2016, Sankaranarayanan2016a} have been proposed to learn from face triplets. The template adaptation \cite{Crosswhite2016} is another strategy which incorporates an SVM classifier and extends the FR tasks for videos and templates \cite{ijbaKlare2015}. A different approach learns an aggregation module \cite{nanyang2016} to compute the similarity scores among two sets of video frames. The principal component analysis (PCA) technique is commonly used \cite{masi16dowe, abdalmageed2016face, masi2016pose} to learn a dataset specific projection matrix. 
The above methods often need to prepare its training data from the target specific datasets. Moreover, they \cite{Sankaranarayanan2016, Sankaranarayanan2016a} may need to carefully prepare the training data, \textit{e.g.}, triplets. \textit{Our FR method do not incorporate any such additional learning strategies}.

FR methods often accumulate features from a set of independently trained CNN models to construct rich facial descriptors, see \emph{\# of CNNs} column of Table \ref{tab:cnn_analysis}. These CNN models are trained with different types of inputs: (a) image-crops based on different facial regions (eyes, nose, lips, \textit{etc}.) \cite{deepid2psun2015, sun2015deepid3, sparsifyingsun2015, wang2016face, ding2015robust}; (b) different images modalities, such as 2D, 3D, frontalized and synthesized faces at different poses \cite{ding2015robust, abdalmageed2016face, masi2016pose, taigman2014deepface} and (c) different training databases with varying number of images \cite{webscaletaigman2015, baiduliu2015}. \textit{Our FR method do not apply these approaches and train single CNN model}.

A large facial image database is significantly important to achieve high accuracy on FR \cite{schroff2015facenet, zhou2015naive, deepvisage_2017_ICCVW}. \textit{Dataset Info} column of Table \ref{tab:cnn_analysis} provides the information of the training datasets used by different FR methods. Currently, several FR datasets \cite{yi2014learning, mscelebguo16} are publicly available. Among them, the CASIA-WebFace \cite{yi2014learning} has been widely used by the recent methods \cite{centerlosswen2016, Sankaranarayanan2016, Sankaranarayanan2016a, viplfacenetliu2016, allinoneranjan2016, yi2014learning, wang2016face, chen2016unconstrained, ding2015robust, wu2015lightened, masi16dowe, abdalmageed2016face, masi2016pose} to train CNN with a medium sized database. Besides, the recently released MSCELEB \cite{mscelebguo16} dataset, which provides the largest collection of facial images and identities, becomes the standard choice for large scale training. \textit{We exploit it for developing our FR method}.
\section{Methodology}
\label{sec:methodology}
In this section, first we present the \textit{statistical feature representation} model and then discuss the facial \textit{feature representation learning} method \textit{w.r.t.} the model. Finally, we present the complete face recognition pipeline.

\subsection{Model and Method}
\label{ssec:model_method}
\subsubsection{Statistical Feature Representation (SFR) Model}
\label{sssec:ffrm}
We propose the SFR model based on the generative model-based approach \cite{murphy2012machine}, where the facial features are issued from a finite statistical mixture of probability distributions. Then, these features are transformed into the 2D image space using a \textit{transformer}. See Appendix \ref{secapp:proof_concept} for the experimental proof of concept of the proposed SFR model with MNIST \cite{lecun1998gradient} digits expements.

Figure \ref{fig:ffr_model}(a) (from right to left) provides an illustration of the SFR model,
which considers a mixture of \textit{von Mises-Fisher} (vMF) distributions \cite{mardia2009directional} to model the facial features from different identities/classes.
The vMF distribution \cite{mardia2009directional} is parameterized with the mean direction $\mu$ (shown as solid lines) and concentration $\kappa$ (indicates the spread of feature points from the solid line). For the $i^{th}$ facial image features $\mathbf{x}_{i}$, called \textit{\textbf{vMF feature}}, we define the SFR model with $M$ classes 
as:
%
\begin{equation}
\label{eq:sffrm}
SFR\left (\mathbf{x}_{i}|\Theta_{M} \right ) = \sum_{j=1}^{M}\pi_{j}V_{d}\left ( \mathbf{x}_{i}|\mu_{j},\kappa_{j} \right )
\end{equation}
where 
$\pi_{j}$, $\mu_{j}$ and $\kappa_{j}$ denote respectively the mixing proportion, mean direction and concentration value of the $j^{th}$ class.
$\Theta_{M}$ is the set of model parameters and $V_{d}(.)$ is the density function of the vMF distribution (see Section \ref{ssec:ffr_vmf_loss} for details).

The SFR model makes \textit{\textbf{equal privilege assumption}} for the classes, \textit{i.e.}, each class $j$ has \textit{\textbf{equal appearance}} probability $\pi$ and is distributed with \textit{\textbf{same concentration}} value $\kappa$. This assumption is important for \textit{discriminative learning} to make sure that the \textit{supervised} classifier is not biased to any particular class regardless of the number of samples and amount of variations present in the training data for each class. On the other hand, $\mu_{j}$ plays significant role to preserve each identity in its respective sub-space. Therefore, the \textit{generative} SFR model can be used for \textit{discriminative} learning tasks, which can be viewed by reversing the directions in Figure \ref{fig:ffr_model}(a), \textit{i.e.}, information will flow from left to right. Next we discuss the discriminative features learning task \textit{w.r.t.} the SFR model in details.
\subsubsection{vMF Feature Learning (vMF-FL) Method}
\label{sssec:frlm}
\begin{figure*}[!t]
\centering
\includegraphics[scale=0.4]{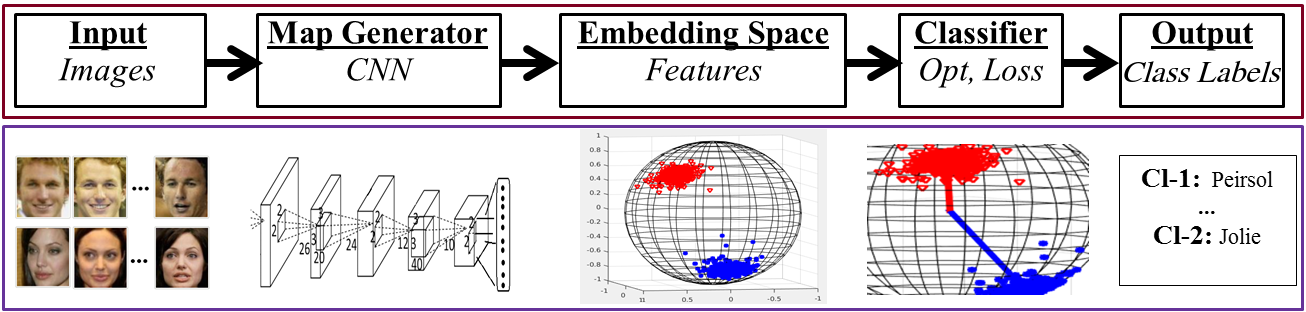}
\caption{Workflow of the vMF-FL method. \textit{\textbf{top:}} block diagram and \textit{\textbf{bottom:}} view with an example.}
\label{fig:frlm}
\end{figure*}
Figure \ref{fig:frlm} illustrates the workflow of the vMF-FL method, where the features are learned using an object identity classifier. See Appendix \ref{secapp:proof_concept} for an extended view of its relationship with the SFR model. The vMF-FL method consists of two sub-tasks: (1) mapping input 2D object images to \textit{vMF feature} using the CNN model, which we use as the \textit{inverse-transformer} w.r.t. the SFR model and (2) classifying features to the respective classes based on the \textit{discriminative} view of the SFR model. It formulates an optimization problem by integrating the SFR and CNN models and learns parameters by minimizing the classification loss. In general, CNN models use the \textit{softmax} function and minimize the cross entropy. Therefore, our integration will replace the \textit{softmax} function according to Eq. \ref{eq:sffrm}. 
\subsubsection{Convolutional Neural Network (CNN) model}
\label{sssec:deep_cnn}
The basic ideas of CNN \cite{lecun1998gradient} consist of: (a) local receptive fields via convolution and (b) spatial sub-sampling via pooling. At layer  $l$, the convolution of the input $f_{x,y}^{Op, l-1}$ to obtain the $k^{th}$ output map $f_{x,y,k}^{C,l}$ is:
\begin{equation}
\label{eq:conv_op}
f_{x,y,k}^{C, l} = {\mathbf{w}_k^l}^T f_{x,y}^{Op, l-1} + b_k^l
\end{equation}
where, $\mathbf{w}_k^l$ and $b_k^l$ are the shared weights and bias. $C$ denotes convolution and $Op$ (for $l>1$) denotes various tasks, such as convolution, sub-sampling or activation. For $l=1$, $Op$ represents the input image.
Sub-sampling or pooling performs a simple local operation, \textit{e.g.}, computing the maximum in a local neighborhood followed by reducing spatial resolution as:
\begin{equation}
\label{eq:max_pool}
f_{x,y,k}^{P, l} = \max_{(m,n) \in \mathcal{N}_{x,y} } f_{m,n,k}^{Op, l-1}
\end{equation}
where, $\mathcal{N}_{x,y}$ is the local neighborhood and $P$ denotes pooling.
In order to ensure non-linearity of the network, the feature maps are passed through a non-linear activation function, \textit{e.g.}, the Rectified Linear Unit (ReLU) \cite{recentcnngu2015, prelu_he2015}: $f_{x,y,k}^l = max(f_{x,y,k}^{l-1}, 0)$.

At the basic level, a CNN is constructed by stacking series of convolution, activation and pooling layers  \cite{lecun1998gradient}. Often a fully connected (FC) layer is placed at the end, which connects all neurons from the previous layer to all neurons of the next layer.
%
%

CNN models are trained by optimizing loss function. The softmax loss, which is widely used for classification, has the following form: 
\begin{equation}
\label{eq:softmax}
\mathcal{L}_{Softmax} = -\sum_{i=1}^{N}log \; \frac{exp(\mathbf{w}^T_{y_i}f_i + b_{y_i})}{\sum_{l=1}^{M}exp(\mathbf{w}^T_lf_i + b_l)}
\end{equation}
where, $f_i$ and $y_i$ are the features and true class label of the $i^{th}$ image. $\mathbf{w}_j$ and $b_j$ denote the weights and bias of the $j^{th}$ class. $N$ and $M$ denote the number of training samples and the number of classes.
\subsection{SFR model and von Mises-Fisher Mixture Loss (vMFML)}
\label{ssec:ffr_vmf_loss}
Our proposed SFR model assumes that the facial features are unit vectors and distributed according to a mixture of vMFs. By combining the SFR and CNN, vMF-ML method provides a novel loss function, called the \textit{von Mises-Fisher Mixture Loss} (vMFML). Below we provide its formulation.
\subsubsection{vMF Mixture Model (vMFMM)}
\label{vmf_distribution}
\begin{figure}
\centering
\includegraphics[scale=0.33]{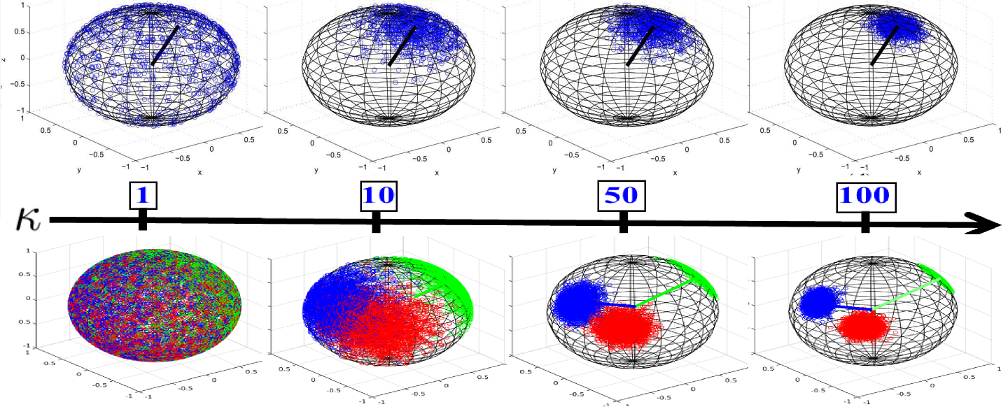}
\caption{3D directional samples from the vMF distribution (above arrow) and the vMFMM of 3 classes (below arrow). Samples are shown on the $S^{2}$ sphere for different values of $\kappa$.}
\label{fig:vmf_dist_mix}
\end{figure}

For a $d$ dimensional random unit vector $\mathbf{x} = \left[x_1,...,x_d \right]^T \in S^{d-1} \subset \R^d$ (\textit{i.e.}, $\left \| \mathbf{x} \right \|_2=1$), the density function of the vMF distribution is defined as \cite{mardia2009directional}:
\begin{equation}
V_{d}(\mathbf{x}|\mu, \kappa)= C_{d}(\kappa)\;\text{exp}(\kappa \mu^{T}\mathbf{x})
\end{equation}
where, $\mu$ denotes the mean (with $\left \| \mu \right \|_2=1$) and $\kappa$ denotes the concentration parameter (with $\kappa\geq 0$). 
$C_{d}(\kappa)=  \frac{\kappa^{d/2-1}}{(2\pi)^{d/2} I_{d/2-1}(\kappa)}$ is the normalization constant, where, $I_{\rho}(.)$ is the modified Bessel function of the first kind.
%
The shape of the vMF distribution depends on the value of the concentration parameter $\kappa$. For high value of $\kappa$, \textit{i.e.} highly concentrated features, the distribution has a mode at the mean direction $\mu$. In contrary, for low values of $\kappa$ the distribution is uniform, \textit{i.e.} the samples appear as to be uniformly distributed on the sphere. 
The top row of Figure \ref{fig:vmf_dist_mix} illustrates examples of the 3D samples in the $S^{2}$ sphere, which are distributed according to the vMF distribution with different values of the concentration $\kappa$.

Let $\mathbf{X} = \left \{ \mathbf{x}_i \right \}_{i=1,...,N}$ is a set of samples, where $N$ is the total number of samples. For the $i^{th}$ sample $\mathbf{x}_i$, the vMFMM with $M$  classes is defined\footnote{Note that, this is equivalent to the definition of SFR model in Eq. \ref{eq:sffrm}} as \cite{banerjee2005clustering}:
%
$g_v\left (\mathbf{x}_{i}|\Theta_{M} \right ) = \sum_{j=1}^{M}\pi_{j}V_{d}\left ( \mathbf{x}_{i}|\mu_{j},\kappa_{j} \right )$,
%
where $\Theta_{M} = \left \{ (\pi_{1},\mu_{1},\kappa_{1}),...,(\pi_{M},\mu_{M},\kappa_{M})\right \}$ is the set of parameters and $\pi_{j}$ is the mixing proportion of the $j^{th}$ class. The bottom row of Figure \ref{fig:vmf_dist_mix} shows the samples from vMFMM with different $\kappa$ values. 

The vMFMM model has been used for \textit{unsupervised classification} \cite{banerjee2005clustering} to cluster data on the unit hyper-sphere, where the model is estimated by the Expectation Maximization (EM) method. The objective of EM method is to estimate model parameters such that the negative log-likelihood value of the vMFMM model, \textit{i.e.}, $-\;log(g_v\left (\mathbf{X}|\Theta_{K} \right ))$, is minimized. The EM method estimates the posterior probability in the \textit{E-step} as \cite{banerjee2005clustering}:
\begin{equation}
\label{eq:posterior_vmf}
p_{ij} = 
\frac{\pi_{j}\; C_{d}(\kappa_j)\;\text{exp}(\kappa_j \mu_j^{T}\mathbf{x}_i)}
{\sum_{l=1}^{K}\pi_{l}\; C_{d}(\kappa_l)\;\text{exp}(\kappa_l \mu_l^{T}\mathbf{x}_i)} 
\end{equation}
and model parameters in the \textit{M-step} as \cite{banerjee2005clustering}: 
\begin{equation}
\begin{multlined}
\label{eq:m_step_vmf}
\pi_{j} = \frac{1}{N}\sum_{i=1}^{N}p_{ij}, \;\; \hat{\mu_{j}} = \frac{\sum_{i=1}^{N}p_{ij} \; \mathbf{x}_i}{\sum_{i=1}^{N}p_{ij}},
\\
 \bar{r} = \frac{\left \| \hat{\mu_{j}} \right \|}{N \; \pi_{j}}, \;\; \mu_j = \frac{\hat{\mu_{j}}}{\left \| \hat{\mu_{j}} \right \|} \;\; \text{and} \;\; \kappa_{j} = \frac{\bar{r} \; d - \bar{r}^3}{1 - \bar{r}^2}
\end{multlined}
\end{equation}
\subsubsection{von Mises-Fisher Mixture Loss (vMFML) and optimization}
\label{ssec:vmfml}
Our vMF-FL method aims to learn discriminative facial features by minimizing the classification loss.
Within this (\textit{supervised classification}) context, we set our objective as to minimize the cross entropy guided by the vMFMM. Therefore, we rewrite 
the posterior probability based on the \textit{\textbf{equal privilege assumption}} of the SFR model as:
\begin{equation}
\label{eq:vmf_posterior_simp}
p_{ij} = 
\frac{\text{exp}(\kappa \mu_j^{T}\mathbf{x}_i)}
{\sum_{l=1}^{M}\text{exp}(\kappa \mu_l^{T}\mathbf{x}_i)}
\end{equation}
Now we can exploit the posterior/conditional probability to minimize the cross entropy and define the loss function, called vMFML, as:
\begin{equation}
\begin{multlined} 
\label{eq:vmfml_loss_soft}
\mathcal{L}_{vMFML} = - \sum_{i=1}^{N} \sum_{j=1}^{M} y_{ij} \; log(p_{ij}) \\ = - \sum_{i=1}^{N} log \frac{\text{exp}(\kappa \mu_j^{T}\mathbf{x}_i)} {\sum_{l=1}^{M}\text{exp}(\kappa \mu_l^{T}\mathbf{x}_i)} \\ = - \sum_{i=1}^{N} log \frac{e^{z_{ij}}} {\sum_{l=1}^{M}e^{z_{il}}}  \; \; \left [z_{ij} = \kappa \mu_j^{T}\mathbf{x}_i  \right ]
\end{multlined}
\end{equation}
where, $y_{ij}$ is the true class probability and we set $y_{ij}=1$ as we only know the true class labels.
%
%
%
%
Now, by comparing the \textit{vMFML} with the \textit{softmax loss} (Eq. \ref{eq:softmax}), we observe the following differences:
(a) vMFML uses unit normalized features: $\mathbf{x} = \frac{\mathbf{f}}{\left \| \mathbf{f} \right \|}$; (b) mean parameter has relation with the softmax weight as: $\mu = \frac{\mathbf{w}}{\left \| \mathbf{w} \right \|}$; (c) it has no bias and (d) it has an additional parameter $\kappa$. 

Now, we observe that the proposed vMF-FL method modifies the CNN training by replacing the \textit{softmax loss} with the  \textit{vMFML}. Therefore, to learn the parameters we can follow the standard CNN model learning procedure, \textit{i.e.}, iteratively learn through the forward and backward propagation \cite{lecun1998gradient}. This requires us to compute the gradients of vMFML \textit{w.r.t.} the parameters. By following the chain rule, we can compute the gradients (we consider single sample and drop subscript $i$ for brevity) as:
\begin{equation}
\label{eq:dpj_dzl}
\frac{\partial \mathcal{L}}{\partial p_j} = -\frac{y_j}{p_j}; 
\frac{\partial p_l}{\partial z_j} = \left\{\begin{matrix}
p_j(1-p_j) & l==j\\ 
-p_j \, p_l & l \neq j 
\end{matrix}\right.; 
\frac{\partial \mathcal{L}}{\partial z_j} = p_j - y_j
\end{equation}
\begin{equation}
\frac{\partial z_j}{\partial \kappa} = \mu_j^{T}\mathbf{x}; \;\;\;\;  
\frac{\partial z_j}{\partial \mu_{jd}} = \kappa \, x_d; \;\;\;\; 
\frac{\partial z_j}{\partial x_d} = \kappa \, \mu_{jd} 
\end{equation}
\begin{equation}
\frac{\partial x_d}{\partial f_d} = \left\{\begin{matrix}
\frac{\partial x_d}{\partial f_d} = \frac{\left \| \mathbf{f} \right \|^2 - f_d^2}{\left \| \mathbf{f} \right \|^3} = \frac{1-x_d^2}{\left \| \mathbf{f} \right \|}\\ 
\frac{\partial x_r}{\partial f_d} = \frac{- f_d f_r}{\left \| \mathbf{f} \right \|^3} = \frac{-x_d x_r}{\left \| \mathbf{f} \right \|}
\end{matrix}\right.
\end{equation}
\begin{equation}
\frac{\partial \mu_d}{\partial w_d} = \left\{\begin{matrix}
\frac{\partial \mu_d}{\partial w_d} = \frac{\left \| \mathbf{w} \right \|^2 - w_d^2}{\left \| \mathbf{w} \right \|^3} = \frac{1-\mu_d^2}{\left \| \mathbf{w} \right \|}\\ 
\frac{\partial \mu_r}{\partial w_d} = \frac{- w_d w_r}{\left \| \mathbf{w} \right \|^3} = \frac{-\mu_d \mu_r}{\left \| \mathbf{w} \right \|}
\end{matrix}\right.
\end{equation}
\begin{equation}
\frac{\partial \mathcal{L}}{\partial \kappa} = \sum_{j=1}^{M} (p_j - y_j) \, \mu_j^T \, \mathbf{x}
\, ; \;\;\;
\frac{\partial \mathcal{L}}{\partial \mu_{jd}} = (p_j - y_j) \, \kappa \, x_d
\end{equation}
\begin{equation}
\label{eq:dloss_df}
\frac{\partial \mathcal{L}}{\partial x_d} = \sum_{j=1}^{M}(p_j - y_j) \, \kappa \, \mu_{jd} ; \;\;\;
\frac{\partial \mathcal{L}}{\partial f_d} = \frac{1}{\left \| \mathbf{f} \right \|} \left ( \frac{\partial \mathcal{L}}{\partial x_d} - x_d \sum_{r} \frac{\partial \mathcal{L}}{\partial x_r} x_r\right )
\end{equation}
\subsection{Interpretation and discussion}
\label{int_discussion}
The proposed SFR model represents each identity/class (\textit{i.e.}, face) with the mean ($\mu$) and concentration ($\kappa$) parameters of the vMF distribution, which (unlike weight and bias) express their direct \textit{relationship} with the respective identity. $\mu$ provides an expected representation (\textit{e.g.}, mean facial features) of the identity and $\kappa$ (independently computed) indicates the variations within the samples from the identity. See Appendix \ref{secapp:proof_concept} for the illustration from MNIST digits based experiment, where the generated images from $\mu_j$ effectively show the ability of vMF-FL to learn a representative images of the corresponding classes.

In terms of \textit{discriminative feature} learning \cite{centerlosswen2016, liu2016large}, we can interpret the effectiveness of vMFML by analyzing the shape of the vMF distributions and vMFMMs in Figure \ref{fig:vmf_dist_mix} based on the $\kappa$ value. While for high $\kappa$ value the features are closely located around the mean direction $\mu$, for low $\kappa$ values they randomly spread around and can be located far from $\mu$.
We observe that $\kappa$ also plays an important role to separate the vMFMM samples from different classes. A higher $\kappa$ value will enforce the features to be more concentrated around $\mu$ to minimize intra-class variations (reduce angular distances of samples and mean) and maximize inter-class distances (see Figure \ref{fig:vmf_dist_mix} and \ref{fig:ffr_model}(b)). Therefore, unlike \cite{centerlosswen2016} (jointly optimizes two losses: \textit{softmax} and \textit{center loss}), we\footnote{See Appendix \ref{secapp:rel_loss_functions} for further details.}
can learn discriminative features by optimizing single loss function and save $M \times D$ parameters, where $D$ is the features dimension. 

The formulation of vMFML can naturally provide interpretation and its relationships with several concurrently proposed loss functions \cite{liu2016large, spherefaceliu2017, l2ranjan2017, normfacewang2017}. In Eq. \ref{eq:vmfml_loss_soft}, by using a higher $\kappa$ value for the true class compared to the rest, \textit{i.e.}, $\kappa_{y_i} > \kappa_{j \neq y_i}$, vMFML can formulate the \textit{large-margin softmax loss }\cite{liu2016large} and \textit{A(angular)-softmax loss }\cite{spherefaceliu2017} under certain conditions\footnote{See Appendix \ref{secapp:rel_loss_functions} for further details.}. 
The \textit{L2-softmax loss }\cite{l2ranjan2017} is similar to vMFML when weights $\left \| \mathbf{w} \right \| = 1$ and biases $b_{y_i}=0$ are applied within its formulation. Indeed, this relationship is significant as it provides additional justification (why $\kappa \mu$ is important) for the vMFML proposal from the perspective of face image quality, see \ref{ssec:disscussion} for further details. The \textit{reformulated softmax loss} proposed in NormFace \cite{normfacewang2017} optimizes the cosine similarity and add a scaling parameter which is empirically found by analyzing the bound of softmax loss. Interestingly, this reformulation (Eq. 6 of \cite{normfacewang2017}) is exactly similar to Eq. \ref{eq:vmfml_loss_soft} and hence vMFML has an interesting interpretation from a different viewpoint. The above discussions clearly state that while vMFML is derived from a theoretical model, it can be explained based on the intuitions and empirical justifications presented in the concurrently proposed loss functions \cite{spherefaceliu2017, l2ranjan2017, normfacewang2017} for FR. 
%
%

\textit{Normalization} \cite{batch_normalization, weight_normalization, cosine_normalization} becomes an increasingly popular technique to use within the CNN models. Our method (\textit{s.t.} normalization in the final layer) takes the advantages of different normalization techniques due to its natural form of the features ($\left \| \mathbf{x} \right \| =1$) and parameter ($\left \| \mu \right \| =1$). Particularly, the term $\kappa \mu$ is equivalent to the reparameterization\footnote{See Appendix \ref{secapp:rel_loss_functions} for further details.} 
proposed by \textit{weight normalization} \cite{weight_normalization} and $\mu^T \mathbf{x}$ is equivalent to the \textit{cosine normalization} \cite{cosine_normalization}. Both \cite{weight_normalization} and \cite{cosine_normalization} provide their relationship with the \textit{batch normalization} \cite{batch_normalization} under certain conditions, which can be equally applicable to our case.
%

\subsection{Face Verification with the vMF-FL method}
\label{ssec:fr_frl_method}
The proposed vMF-FL method learns discriminative facial feature representation from a set of 2D facial images. Therefore, we use it to extract facial features and verify pairs of face images \cite{lfw_huang2014}, templates \cite{ijbaKlare2015} and videos \cite{ytfwolf2011}. 
%
\subsubsection{CNN model}
\label{proposed_cnn_model}
\begin{figure}[!t]
\centering
\subfloat[]{\includegraphics[scale=0.18]{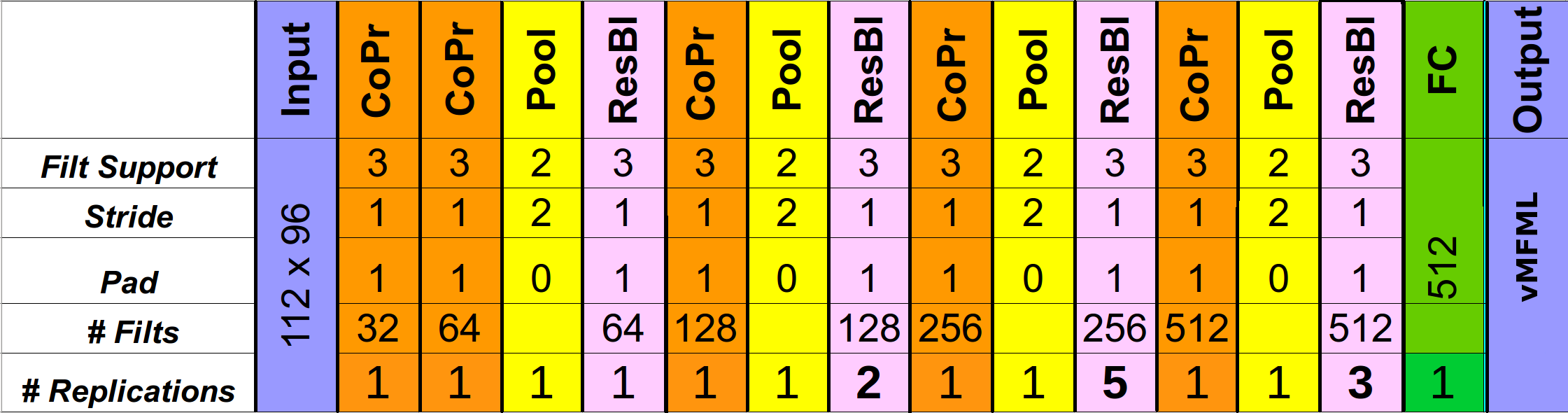}}
\hfill
\subfloat[]{\includegraphics[scale=0.25]{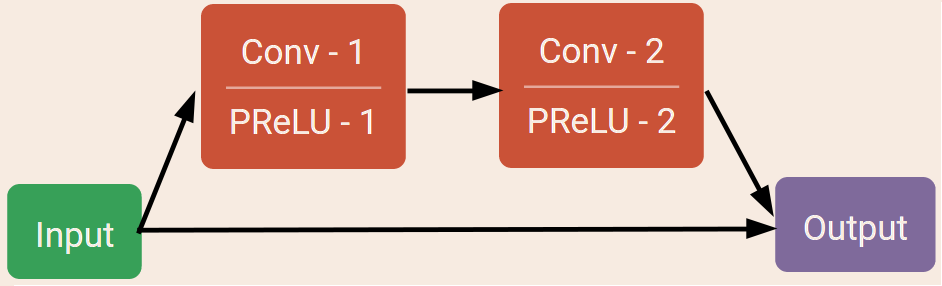}}
\caption{(a) Illustration of the CNN model with vMFML. \textbf{\textit{CoPr}} indicates \textit{Convolution} followed by the \textit{PReLU} activation. \textbf{\textit{ResBl}} is a residual block \cite{resnethe2015deep} which computes $output \; = \; input + CoPr(CoPr(input))$. \textit{\textbf{\# Replication}} indicates how many times the same block is sequentially replicated. \textit{\textbf{\# Filts}} denotes the number of feature maps. (b) illustration of the residual block \textbf{\textit{ResBl}}.}
\label{fig:res_block_cnn}
\end{figure}

In general, any CNN model can be used with the proposed vMF-FL method. In this work, we follow the recent trend  \cite{recentcnngu2015, resnethe2015deep} and use a deeper CNN. To this aim, we choose the publicly available\footnote{Note that the CNN proposed in \cite{centerlosswen2016} is different than the publicly provided CNN by the same authors. Therefore, in order to avoid confusion, we do not cite our CNN model directly as \cite{centerlosswen2016}.} ResNet \cite{resnethe2015deep} based CNN model provided by the authors of \cite{centerlosswen2016}. 
It consists of 27 convolution (\textit{Conv}), 4 pooling (\textit{Pool}) and 1 fully connected (\textit{FC}) layers. 
Figure \ref{fig:res_block_cnn} illustrates the CNN model, called \textbf{\textit{Res-27}}.
Each convolution uses a $3\times3$ kernel and is followed by a PReLU activation function. The CNN progresses from the lower to higher depth by decreasing the spatial resolution using a $2\times2$ \textit{max Pool} layer while gradually increasing the number of feature maps from 32 to 512. 
The 512 dimensional output from the FC layer is then unit normalized which we consider as the desired directional features representation of the input 2D image. Finally, we use the proposed vMFML and optimize the CNN during training. Overall, the CNN comprises 36.1M parameters for feature representation and $(512 \times M)+1$ parameters for the vMFML, where M is the total number of identities/classes in the training database. Note that, vMFML only requires one additional scalar parameter ($\kappa$) compared to the general softmax loss.
\subsubsection{Face verification}
\label{par:face_ver_test}
Our face verification strategy follows the steps below:
\begin{enumerate}
\item
\textit{\textbf{face image normalization}}: we apply the following steps: (a) use the MTCNN \cite{mtcnnZhang2016} detector\footnote{In case of multiple faces, we take the face closer to the image center. If the landmarks detector fails we keep the face image by cropping it based on the detected bounding box.} to detect the faces and landmarks; (b) 
normalize the face image by applying a 2D similarity transformation. The transformation parameters are computed from the location of the detected landmarks on the image and pre-set coordinates in a 112$\times$96 image frame; (c) apply color to grayscale image conversion and (d) normalize the pixel intensity values within the range $-1$ to $1$.
Note that, this normalization method is also applied during the training data preparation.
\item
\textit{\textbf{features extraction}}: use the CNN (trained with vMF-FL) to extract features from the original and horizontally flipped version and take the element-wise maximum value. For template \cite{ijbaKlare2015} and video \cite{ytfwolf2011}, obtain the features of an identity by taking element-wise average of the features from all of the images/frames.
\item
\textit{\textbf{verification score computation}}: compute the cosine similarity as the score and compare it to a threshold.
\end{enumerate}

\section{Experiments, Results and Discussion}
\label{sec:res_exp}
We train the CNN model, use it to extract features and perform different scenarios for face verification, \textit{i.e.}, single image-based \cite{lfw_huang2014, chen2015facecacd}, multi-image or video-based \cite{ijbaKlare2015, ytfwolf2011}. In order to verify the effectiveness of the proposed approach, we experiment on several datasets, namely LFW \cite{lfw_huang2014}, IJB-A \cite{ijbaKlare2015}, YTF \cite{ytfwolf2011} and CACD \cite{chen2015facecacd},
which impose various challenge to FR by collecting images from different sources and ensuring sufficient variations \textit{w.r.t.} pose, illumination, occlusion, expression, resolution, age, geographic regions, \textit{etc}.
Figure \ref{fig:incorrect_dbs} provides examples of the pairs of images/videos/templates from different FR datasets.

LFW \cite{lfw_huang2014} dataset was particularly designed to study the problem of face verification with unconstrained facial images taken in everyday settings and collected from the web. It is considered as one of the most challenging and popular FR benchmark. YTF \cite{ytfwolf2011} is a dataset of unconstrained videos and designed to evaluate FR methods for matching faces in pairs of videos. Low image resolution, high illumination and pose variations on the YTF videos added significant challenges to the video based FR methods. IJB-A \cite{ijbaKlare2015} is a recently proposed dataset, which raises the FR difficulty by collecting faces with high variations in pose, illumination, expression, resolution and occlusion. Unlike LFW, IJB-A includes faces which cannot be detected by standard face detectors and hence added significant amount of challenges. The CACD \cite{chen2015facecacd} dataset is specifically designed to study age invariant FR problem. It ensures large variations of the ages in order to add further challenge for the FR methods.
\subsection{CNN Training}
\label{ssec:cnn_train}
We collect the training images from the cleaned\footnote{We take the list of 5.05M faces provided by \cite{wu2015lightened} and keep non-overlapping (with test set) identities which have at least 30 images after successful landmarks detection.} version of the MS-Celeb-1M \cite{mscelebguo16} database, which consists of 4.61M images of 61.24K identities. In order to pre-process the training data we normalize (Sect. \ref{par:face_ver_test}) the facial images of the dataset.

We train our CNN model using only the identity label of each image. We use 95\% images (4.3M images) for training and 5\% images (259K images) for monitoring and evaluating the loss and accuracy. We train our CNN using the \textit{stochastic gradient descent} method and \textit{momentum} set to 0.9. 
Moreover, we apply \textit{L2} regularization with the \textit{weight decay} set to $5e^{-4}$. 
We begin the CNN training with a learning rate 0.1 for 2 epochs. 
Then we decrease it after each epoch by a factor 10. We stop the training after 5 epochs.
We use 120 images in each mini-batch. During training, we apply data augmentation by horizontally flipping the images. Note that, during evaluation on a particular dataset, we do not apply any additional CNN training or fine-tuning or dimension reduction.
%
%
%
%
\subsection{Results and Evaluation}
\label{ssec:res_eval}
First we evaluate the proposed \textit{vMFML} by comparing it to the state-of-the-art loss functions. Next, we evaluate our \textit{vMF-FL} based FR method on the most common and challenging FR benchmark datasets and compare it to the state-of-the-art FR methods.
\subsubsection{Comparison of the Loss Functions}
\label{sss:comp_loss_func}
To gain insight on the effectiveness of the proposed loss function,  we compare the proposed vMFML with some state-of-the-art loss functions using two different CNN architectures of different depth along with two training datasets of different size. Table \ref{tab:sensitivity_comp_loss} presents the results, where we use the LFW \cite{lfw_huang2014} face verification accuracy as a measure of the performance.
We consider three commonly used loss functions\footnote{Note that, we do not compare with the contrastive loss \cite{contrastive_loss} as the joint \textit{softmax+center} loss (JSCL) \cite{centerlosswen2016} has been shown to be more efficient than it.} for FR, such as, Softmax loss, joint \textit{softmax+center} loss (JSCL) \cite{centerlosswen2016} and \textit{softmax followed by triplet} loss (STL) \cite{schroff2015facenet}.
Note that, we use the same CNN model and best known training settings (learning rate, regularization, \textit{etc}.) for individual loss function during the experiments.

\begin{figure}[t]
\centering
\includegraphics[scale=0.4]{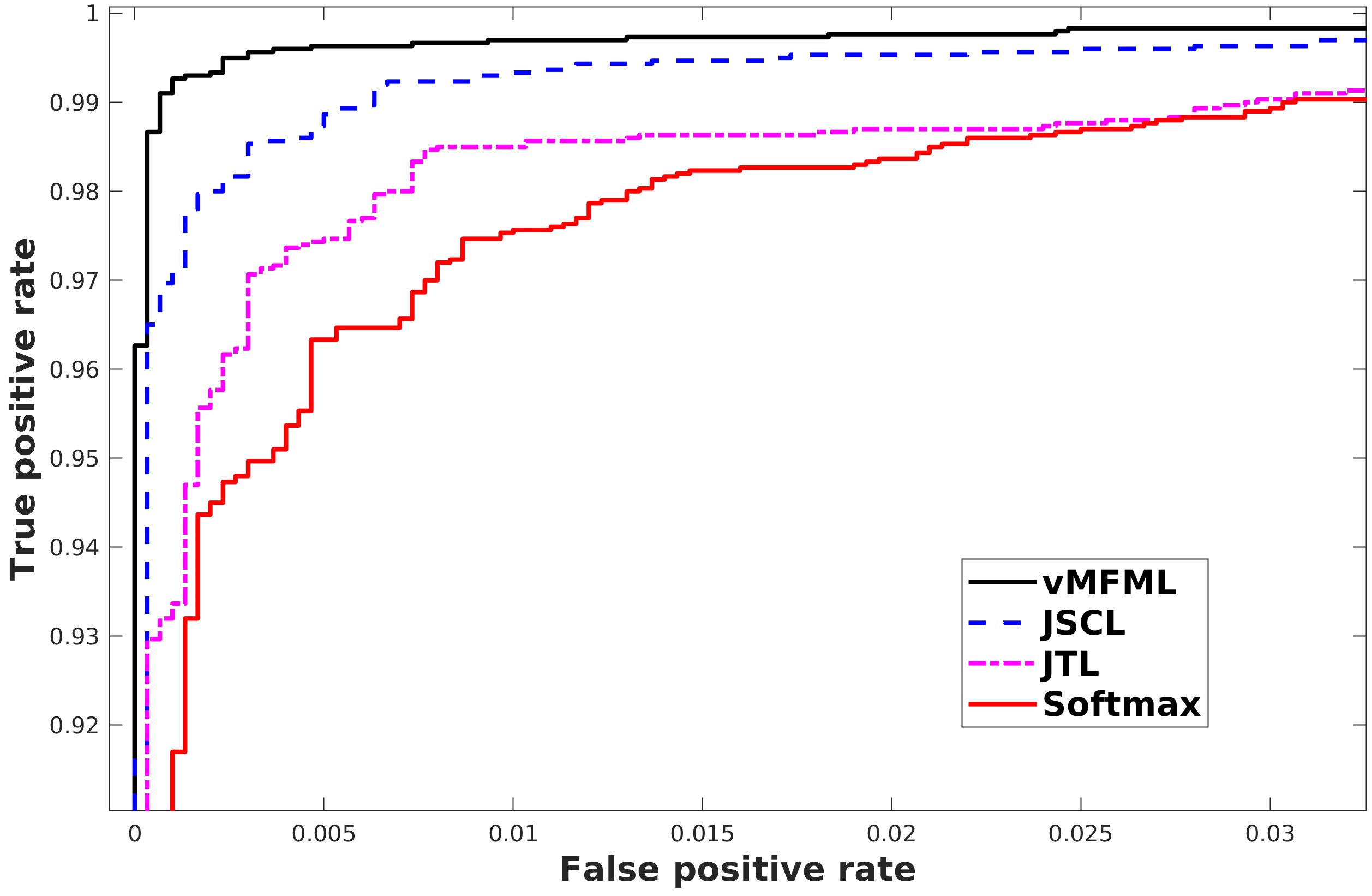}
\caption{\footnotesize{Illustration of the ROC curves for different loss functions: vMF Mixture Loss (vMFML), Softmax loss, joint \textit{softmax+center} loss (JSCL) \cite{centerlosswen2016} and \textit{softmax followed by triplet} loss (STL) \cite{schroff2015facenet}. Results obtained using the same CNN settings trained with the MS-Celeb-1M \cite{mscelebguo16} dataset.}}
\label{fig:roc_det_plot}
\end{figure}

First, we use the CASIA dataset \cite{yi2014learning} and train a shallower CNN proposed in \cite{yi2014learning}, called \textit{CasiaNet}. We observe that:
vMFML (98\% ) $>$ STL (97.93\%) $>$ JSCL (97.6\%) $>$ softmax (97.5\%).
Next, we train a deeper CNN, called \textit{Res-27} (Section \ref{proposed_cnn_model}), with the same dataset and observe that:
vMFML (99.18\% ) $>$ JSCL (98.87\%) $>$ STL (98.13\%) $>$ softmax (97.4\%).
Therefore, \textit{vMFML achieves better result (from 98\% to 99.18\%) with a deeper CNN}. Fig. \ref{fig:roc_det_plot} illustrates the corresponding receiver operating characteristic (ROC) curve for this comparison. Next, we train \textit{Res-27} with the MS-Celeb-1M \cite{mscelebguo16} and observe that:
vMFML (99.63\% ) $>$ JSCL (99.28\%) $>$ STL (98.83\%) $>$ Softmax (98.50\%).
That means, \textit{vMFML improves its accuracy (from 99.18\% to 99.63\%) when trained with a larger dataset}. We also analyze the influence of  a larger dataset by training \textit{CasiaNet} with the MS-Celeb-1M \cite{mscelebguo16} and observe that: vMFML (98.65\% ) $>$ STL (98.41\%) $>$ JSCL (98.33\%) $>$ Softmax (98.21\%). This observation further verifies the importance of a larger dataset to achieve better results with deep CNN, including  the proposed vMFML.

The above analyses indicate that: 1) the \textit{proposed vMFML} outperforms the state-of-the-art loss functions commonly used for FR; 2) \textit{Res-27} is an appropriate CNN model and 3) the MS-Celeb-1M \cite{mscelebguo16} is the best choice of training dataset to develop our FR method (see Sect. \ref{ssec:fr_frl_method} and \ref{ssec:cnn_train}). In the next Section, we compare our FR method with the state-of-the-art for a variety of face verification tasks using different benchmarks. Besides, we conduct additional experiments and analysis to identify several common issues related to CNN training and discuss them in Sect. \ref{ssec:disscussion} .

%
%
%
\begin{table}[]
\centering
\caption{Comparison of different loss functions evaluated \textit{w.r.t.} different CNN architectures and training datasets. The LFW \cite{lfw_huang2014} verification accuracy is used as a measure of the performance. Loss functions: vMF Mixture Loss (vMFML), Softmax loss, joint \textit{softmax+center} loss (JSCL) \cite{centerlosswen2016} and \textit{softmax followed by triplet} loss (STL) \cite{schroff2015facenet}.}
\label{tab:sensitivity_comp_loss}
\begin{tabular}{|c|c|c|c|c|c|}
\hline
\textbf{Database} & \textbf{CNN} & \textbf{Softmax} & \textbf{JSCL} & \textbf{STL} & \textbf{vMFML} \\ \hline
Casia             & CasiaNet     & 97.50            & 97.60         & 97.93        & 98.00          \\ \hline
Casia             & Res-27       & 97.40            & 98.87         & 98.13        & 99.18           \\ \hline
MSCeleb-1M        & Res-27       & 98.50            & 99.28         & 98.83        & 99.63          \\ \hline
MSCeleb-1M        & CasiaNet     & 98.21            & 98.33         & 98.42        & 98.65          \\ \hline
\end{tabular}
\end{table}
%


\subsubsection{Labeled Faces in the Wild (LFW)} 
\label{sss:lfw_eval}
The \textit{LFW} dataset \cite{lfw_huang2014} is the \textit{de facto benchmark} for evaluating unconstrained FR methods based on single image comparison. It consists of 13,233 images from 5,759 identities. Fig. \ref{fig:incorrect_dbs}(a) illustrates some face images from this dataset.
The FR task requires verifying 6000 image pairs, which are equally divided into genuine and impostor pairs and comprises total 7.7K images from 4,281 identities. The LFW evaluation requires face verification in 10 pre-specified folds and report the average accuracy. It has different evaluation protocols and based on the recent trend we follow the \textit{unrestricted-labeled-outside-data} protocol. Table \ref{tab:lfw_comparison} presents the results from our method in comparison with other state-of-the-art methods.

As can be seen in Table \ref{tab:lfw_comparison}, our method achieves very competitive accuracy (99.63\%) and ranks among the top performers, 
despite the fact that: (a) L2-Softmax \cite{l2ranjan2017} used\footnote{They achieved 99.60\% when using the same CNN model that we used.} a 100 layers  CNN model (3.7 times deeper/larger compared to us) to achieve 99.78\%; (b) Baidu \cite{baiduliu2015} combines 10 CNNs to obtain 99.77\%, whereas we use a single CNN and (c) FaceNet \cite{schroff2015facenet} used 200M images of 8M identities to obtain  99.63\%, whereas we train our CNN with only 4.51M images and 61.24K identities.
It is interesting to note that, our LFW result (99.63\%) is obtained by training CNN with the cleaned  MSCeleb dataset that removes LFW overlapped identities. In removing the incorrectly  labeled pairs within LFW (see Errata on the LFW webpage), our proposed vMF-FL further displays  99.68\% accuracy.  
In light of the results from \cite{baiduliu2015, sparsifyingsun2015, sun2015deepid3, zhou2015naive, aifrwen2016latent, deepid2psun2015, l2ranjan2017}, the performance of the proposed method could be further improved   by combining features from multiple CNN models or by using deeper (\textit{e.g.}, 100 layers) CNNs.
\begin{table}[h]
\footnotesize
\centering
\caption{\footnotesize{Comparison of the state-of-the-art methods evaluated on the LFW benchmark \cite{lfw_huang2014}.}}
\label{tab:lfw_comparison}
\begin{tabular}{|c|c|c|c|}
\hline
\textbf{FR method} & \textbf{\begin{tabular}[c]{@{}c@{}}\# of\\ CNNs\end{tabular}} & \textbf{\begin{tabular}[c]{@{}c@{}}Dataset\\ Info\end{tabular}} & \textbf{\begin{tabular}[c]{@{}c@{}}Acc\\ \%\end{tabular}} \\ \hline
\textit{\textbf{vMF-FL (proposed)}} & 1 & 4.51M, 61.24K & 99.63 \\ \hline
Baidu \cite{baiduliu2015}        & 10 & 1.2M, 1.8K & 99.77 \\ \hline
Baidu \cite{baiduliu2015}        & 1 & 1.2M, 1.8K & 99.13 \\ \hline
L2-Softmax \cite{l2ranjan2017} & 1 & 3.7, 58.2K & 99.78 \\ \hline
FaceNet \cite{schroff2015facenet}        & 1 & 200M, 8M & 99.63 \\ \hline
DeepVisage \cite{deepvisage_2017_ICCVW} & 1 & 4.48M, 62K & 99.62 \\ \hline
RangeLoss \cite{rangelosszhang2016}        & 1 & 1.5M, 100K & 99.52 \\ \hline
Sparse ConvNet \cite{sparsifyingsun2015} & 25  & 0.29M, 12K & 99.55 \\ \hline
DeepID3 \cite{sun2015deepid3}            & 25  & 0.29M, 12K & 99.53 \\ \hline
Megvii \cite{zhou2015naive}              & 4 & 5M, 0.2M & 99.50 \\ \hline
LF-CNNs \cite{aifrwen2016latent}     & 25 & 0.7M, 17.2K & 99.50 \\ \hline
DeepID2+ \cite{deepid2psun2015}            & 25  & 0.29M, 12K & 99.47 \\ \hline
SphereFace \cite{spherefaceliu2017} & 1  & 0.49M, 10.57K & 99.42 \\ \hline
Center Loss \cite{centerlosswen2016}     & 1 & 0.7M, 17.2K & 99.28 \\ \hline
NormFace \cite{normfacewang2017} 			& 1 & 0.49M, 10.5K & 99.19 \\ \hline
MM-DFR \cite{ding2015robust}             & 8 & 0.49M, 10.57K & 99.02 \\ \hline
VGG Face \cite{parkhi2015deep}           & 1 & 2.6M, 2.6K & 98.95 \\ \hline
MFM-CNN \cite{wu2015lightened}         & 1 & 5.1M, 79K & 98.80 \\ \hline
VIPLFaceNet	\cite{viplfacenetliu2016}   & 1 & 0.49M, 10.57K & 98.60 \\ \hline
Webscale \cite{webscaletaigman2015}     & 4 & 4.5M, 55K  & 98.37 \\ \hline
AAL \cite{aalye2016face}				   & 1 & 0.49M, 10.57K & 98.30 \\ \hline
FSS \cite{wang2016face}		   & 9 & 0.49M, 10.57K & 98.20 \\ \hline
Face-Aug-Pose-Syn \cite{masi16dowe} 			& 1 & 2.4M, 10.57K & 98.06 \\ \hline
CASIA-Webface \cite{yi2014learning}      & 1 & 0.49M, 10.57K & 97.73 \\ \hline
Unconstrained FV \cite{chen2016unconstrained} & 1 & 0.49M, 10.5K & 97.45 \\ \hline
Deepface \cite{taigman2014deepface}      & 3 & 4.4M, 4K & 97.35 \\ \hline
\end{tabular}
\end{table}
However, Table \ref{tab:lfw_comparison} suggests saturation in the LFW results, as most of the methods already surpass human performance (97.53\%) and they differ very few in terms of performance. Besides, it raises debate whether we should justify a method \textit{w.r.t.} the real world FR scenario \cite{blufrliao2014} based on the matching of only 6K pairs. To overcome this limitation, we follow the BLUFR LFW protocol \cite{blufrliao2014}, which exploits all LFW images, verifies 47M pairs per trial and provide the measure of true accept rate (TAR) at a low false accept rate (FAR). In Table \ref{blufr_comp}, we provide our result for the verification rate (VR) at FAR=0.1\% and compare it with the other methods which reported results in this protocol. As can be seen in Table \ref{blufr_comp}, our method displays the best performance in comparison with  the results of some state of the FR methods published so far. 
%
%
%
%
\begin{table}[]
\centering
\caption{Comparison of the state-of-the-art methods based on the BLUFR LFW protocol \cite{blufrliao2014}.}
\label{blufr_comp}
\begin{tabular}{|c|c|}
\hline
\textbf{Method}                        & \textbf{VR@FAR=0.1\%} \\ \hline
\textbf{vMF-FL(proposed)}              & 99.10                 \\ \hline
DeepVisage \cite{deepvisage_2017_ICCVW} & 98.65				\\ \hline
NormFace \cite{normfacewang2017}     & 95.83                 \\ \hline
Center Loss \cite{centerlosswen2016} & 92.97                 \\ \hline
FSS \cite{wang2016face}              & 89.80                 \\ \hline
CASIA \cite{yi2014learning}          & 80.26                 \\ \hline
\end{tabular}
\end{table}

Besides evaluating with different performance measures, we further challenge our proposed FR method  on LFW from the age invariant perspective and benchmark it on the recently released variant of LFW, called Cross-Age LFW (CALFW) dataset \cite{calfwzheng2017}. CALFW is constructed by re-organizing (by crowdsourcing efforts) the LFW verification pairs with apparent age gaps (as large as possible) to form the positive pairs and then selecting negative pairs using individuals with the same race and gender (see Fig. \ref{fig:incorrect_dbs}(b) for some illustrations). Similar to LFW, CALFW evaluation consists of verifying 6000 pairs (equally divided into genuine and impostor pairs) of images in 10 pre-specified folds and report the average accuracy. Table \ref{calfw_comp} synthesizes the experimental result in comparison with the state of the art. As can be seen from Table \ref{calfw_comp},   our method provides the best results compared to the other available results from different methods. 

\begin{table}[!t]
\centering
\caption{Evaluation of the CALFW \cite{calfwzheng2017} dataset. Results from competitive methods are obtained from the paper \cite{calfwzheng2017}.}
\label{calfw_comp}
\begin{tabular}{|c|c|c|c|}
\hline
             & \textbf{\textit{vMF-FL}} \textit{\textbf{(proposed)}} & VGG-Face  & Noisy Softmax\\ \hline
\textbf{Acc\%} & \textbf{94.20} & 86.50 & 82.52 \\ \hline
\end{tabular}
\end{table}

The results in Table \ref{tab:lfw_comparison}, \ref{blufr_comp}, \ref{calfw_comp} confirm the remarkable performance of \textit{vMF-FL} on the LFW dataset. Next, we explore the FR task on videos and evaluate our method on the YouTube Faces \cite{ytfwolf2011} dataset.
\subsubsection{YouTube Faces (YTF) }
\label{sss:ytf_eval}
%
The \textit{YTF} database \cite{ytfwolf2011} (see Fig. \ref{fig:incorrect_dbs}(c) for some illustrations) is the \textit{de facto benchmark} for evaluating the FR methods based on the unconstrained videos. 
It consists of 3,425 videos from 1,595 identities. Evaluation with YTF requires matching 5000 video pairs in 10 folds experiment and report the average accuracy. Each fold consists of 500 video pairs and ensures subject-mutually exclusive property. We follow the \textit{restricted} protocol of YTF which restricts the access to only the similarity information. Table \ref{tab:ytf_comparison} provides the result from our method and compare it with the state-of-the-art methods.

Results in Table \ref{tab:ytf_comparison} show that our method provides the best accuracy (96.46\%) in this dataset for the \textit{restricted} protocol.
In Table \ref{tab:ytf_comparison}, we also report the results (separated with a horizontal line) from the \textit{unrestricted} protocol, \textit{i.e.}, access to similarity and identity \footnote{Access to identity information helps to create a large number of similar and dissimilar image pairs and hence can be further used to train or fine-tune CNN with an additional loss function, \textit{e.g.}, the contrastive or the triplet loss \cite{parkhi2015deep}.} 
information of the test data. 
Comparison indicates that our method is very competitive to the results from \textit{unrestricted} protocol also. Moreover, by comparing the VGG Face \cite{parkhi2015deep} results with both protocols we observe that its accuracy increases significantly (from \textit{restricted}-91.6\% to \textit{unrestricted}-97.3\%) when the CNN is fine-tuned on the YTF dataset. Therefore, this comparison indicates that our result (96.46\%) can be further enhanced if we train or fine-tune our CNN using the YTF data.

%
%
\begin{table}[!t]
\centering
\caption{\footnotesize{Comparison of the state-of-the-art methods evaluated on the Youtube Face \cite{ytfwolf2011}. \textit{\textbf{Ad.Tr.}} denotes additional training is used.}}
\label{tab:ytf_comparison}
\begin{tabular}{|c|c|c|}
\hline
\textbf{FR method} & \textbf{\begin{tabular}[c]{@{}c@{}}Ad.Tr.\end{tabular}} & \textbf{\begin{tabular}[c]{@{}c@{}}Accuracy (\%)\end{tabular}} \\ \hline
\textbf{\textit{vMF-FL (proposed)}}  & N & 96.46\\ \hline
DeepVisage \cite{deepvisage_2017_ICCVW} & N & 96.34\\ \hline
L2-Softmax \cite{l2ranjan2017}	& N & 96.08\\ \hline
VGG Face \cite{parkhi2015deep}           & N & 91.60\\ \hline
Sparse ConvNet \cite{sparsifyingsun2015} & N & 93.50\\ \hline
FaceNet \cite{schroff2015facenet}        & N & 95.18 \\ \hline
SphereFace \cite{spherefaceliu2017} 		& N  & 95.00 \\ \hline
Center Loss \cite{centerlosswen2016}     & N & 94.90 \\ \hline
NormFace \cite{normfacewang2017} 			& Y & 94.72 \\ \hline
RangeLoss \cite{rangelosszhang2016}        & N & 93.7 \\ \hline
DeepID2+ \cite{deepid2psun2015}          & N & 93.20 \\ \hline
MFM-CNN \cite{wu2015lightened}         & N & 93.40 \\ \hline
CASIA-Webface \cite{yi2014learning}      & Y & 92.24 \\ \hline
Deepface \cite{taigman2014deepface}      & Y & 91.40 \\ \hline
\hline \hline
VGG Face \cite{parkhi2015deep}           & Y & 97.30\\ \hline
NAN \cite{nanyang2016} 				    & Y  & 95.72 \\ \hline
\end{tabular}
\end{table}
\subsubsection{Cross-Age Celebrity Dataset (CACD) }
\label{sss:cacd_eval}
%
The \textit{CACD} dataset \cite{chen2015facecacd} aims to incorporate additional challenges to the FR task by explicitly focusing on the age invariant scenario (see Fig. \ref{fig:incorrect_dbs}(d) for some illustrations). Therefore, it ensures a large variations of the ages within the collected database of faces in the wild. 
CACD is constructed by collecting 163,446 images from 2000 identities, where the range of ages varies from 16 to 62. FR with the CACD dataset requires evaluating the similarity of 4000 pairs of images, which are equally splitted in ten folds experimental settings. The average from these ten folds accuracy is considered as the measure of evaluation. In Table \ref{tab:cacd_comparison}, we provide the results of \textit{vMF-FL} and compare it with the state-of-the-art methods.

Results from Table \ref{tab:cacd_comparison} show that our method provides the best accuracy. Moreover, it is better than LF-CNN \cite{aifrwen2016latent}, which is a recent method specialized on age invariant face recognition.
\begin{table}[!t]
\centering
\caption{\footnotesize{Comparison of the state-of-the-art methods evaluated on the CACD \cite{chen2015facecacd} dataset. VGG \cite{parkhi2015deep} result is obtained from \cite{wu2015lightened}.}}
\label{tab:cacd_comparison}
\begin{tabular}{|c|c|c|}
\hline
\textbf{FR method} & \textbf{\begin{tabular}[c]{@{}c@{}}Accuracy (\%)\end{tabular}} \\ \hline
\textbf{\textit{vMF-FL (proposed)}}  & 99.20\\ \hline
DeepVisage \cite{deepvisage_2017_ICCVW} & 99.13 \\ \hline
LF-CNNs \cite{aifrwen2016latent}         & 98.50 \\ \hline
MFM-CNN \cite{wu2015lightened}           & 97.95 \\ \hline
VGG Face \cite{parkhi2015deep}           & 96.00\\ \hline
CARC \cite{chen2015facecacd}             & 87.60 \\ \hline \hline
Human, Avg.             				   & 85.70 \\ \hline
Human, Voting \cite{chen2015facecacd}    & 94.20 \\ \hline
\end{tabular}
\end{table}
\subsubsection{IARPA Janus Benchmark A (IJB-A) }
\label{sss:ijba_eval}
The recently proposed \textit{IJB-A} database \cite{ijbaKlare2015} is developed with the aim to augment more challenges to the FR task by collecting facial images with a wide variations in pose, illumination, expression, resolution and occlusion (see Fig. \ref{fig:incorrect_dbs}(e) for some illustrations). 
IJB-A is constructed by collecting 5,712 images and 2,085 videos from 500 identities, with an average of 11.4 images and 4.2 videos per identity. We follow the \textit{compare} protocol of IJB-A because it measures the face verification accuracy. This protocol requires the comparison among pairs of templates, where each template consists of a set of images and video-frames. The evaluation protocol of IJB-A computes the true accept rate (TAR) at different fixed false accept rate (FAR), \textit{e.g.}, 0.01 and 0.001. The evaluation requires computing the metrics in ten random split experimental settings. Note that we do not use the training data from each split and only evaluate on the test data using our once-trained FR method.

In Table \ref{tab:ijb_comparison}, we present the results of \textit{vMF-FL} and compare it with the state-of-the-art methods. We separate the results (with a horizontal line) to distinguish two categories of methods: (1) methods do not use IJB-A training data and evaluate test sets with a pre-trained CNN; our method belongs to this category and (2) methods use IJB-A training data and applies additional training, such as, CNN fine-tuning or metric learning.

From the results in Table \ref{tab:ijb_comparison}, we observe that, while L2-softmax \cite{l2ranjan2017} provides the best result, our method provides competitive results among the others. We manually investigated our performance on IJB-A and observed that a large number of images were incorrectly pre-processed because of the failure of face and landmarks detection performance by the pre-processor. This is due to the fact that a large part of this dataset consists of very high pose and occluded faces, see Fig. \ref{fig:incorrect_dbs}(e) for few examples. 
Therefore, we conjecture that our performance could be much more improved if we apply a better pre-processor for this challenging database.

Note that, there are numerous methods, such as TA \cite{Crosswhite2016}, NAN \cite{nanyang2016} and TPE \cite{Sankaranarayanan2016}, which use the CNN features and incorporates additional learning method to improve the results. Therefore, features from vMF-FL can be used with them \cite{Crosswhite2016, nanyang2016, Sankaranarayanan2016} to further improve the results.
\subsubsection{Summary of the experimental results }
\label{sss:Summary of the experimental results }
Results of \textit{vMF-FL} on different datasets prove that besides achieving significant results it generalizes very well and overcomes several difficulties which make unconstrained FR a challenging task.
\begin{table}[t]
\footnotesize
\centering
\caption{\footnotesize{Comparison of the state-of-the-art methods evaluated on the IJB-A benchmark \cite{ijbaKlare2015}. `-' indicates the information for the entry is unavailable. Methods which incorporates external training (ExTr) or CNN fine-tuning (FT) with IJB-A training data are separated with a horizontal line. VGG-Face result was provided by \cite{Sankaranarayanan2016}. T@F denotes the \textit{True Accept Rate at a	fixed False Accept Rate (TAR@FAR)}.}}
\label{tab:ijb_comparison}
\begin{tabular}{|c|c|c|c|c|}
\hline
\textbf{FR method} & \textbf{\begin{tabular}[c]{@{}c@{}}ExTr\end{tabular}} & \textbf{\begin{tabular}[c]{@{}c@{}}FT\end{tabular}} & \textbf{\begin{tabular}[c]{@{}c@{}}T@F\\0.01\end{tabular}} & \textbf{\begin{tabular}[c]{@{}c@{}}T@F\\0.001\end{tabular}} \\ \hline
\textit{\textbf{vMF-FL (proposed)}}  & N & N & 0.897 & 0.850\\ \hline
L2-softmax \cite{l2ranjan2017} & N & N & 0.968 & 0.938\\ \hline
DeepVisage \cite{deepvisage_2017_ICCVW} & N & N & 0.887 & 0.824\\ \hline
VGG Face \cite{parkhi2015deep}           & N & N & 0.805 & 0.604\\ \hline
Face-Aug-Pose-Syn \cite{masi16dowe} 			& N & N & 0.886 & 0.725\\ \hline
Deep Multipose \cite{abdalmageed2016face} & N & N  & 0.787 & - \\ \hline
Pose aware FR \cite{masi2016pose} & N & N  & 0.826 & 0.652 \\ \hline
TPE \cite{Sankaranarayanan2016a} & N & N  & 0.871 & 0.766\\ \hline
All-In-One \cite{allinoneranjan2016} & N & N  & 0.893 & 0.787 \\ \hline \hline \hline
L2-softmax \cite{l2ranjan2017} & Y & N & 0.970 & 0.943\\ \hline
All-In-One \cite{allinoneranjan2016} + TPE & Y & N  & 0.922 & 0.823 \\ \hline
Sparse ConvNet \cite{sparsifyingsun2015} & Y & N  & 0.726 & 0.460\\ \hline
FSS \cite{wang2016face}		   & N & Y  & 0.729 & 0.510\\ \hline
TPE \cite{Sankaranarayanan2016a} & Y & N  & 0.900 & 0.813\\ \hline
Unconstrained FV \cite{chen2016unconstrained} & Y & Y  & 0.838 & -\\ \hline
TSE \cite{Sankaranarayanan2016} & Y & Y  & 0.790 & 0.590\\ \hline
NAN \cite{nanyang2016} & Y & N  & 0.941 & 0.881 \\ \hline
TA \cite{Crosswhite2016} & Y & N  & 0.939 & 0.836 \\ \hline
End-To-End \cite{eteChen2015} & N & Y  & 0.787 & - \\ \hline
\end{tabular}
\end{table}
\subsection{Additional Analysis and Discussion}
\label{ssec:disscussion}
In this section, we perform in-depth analysis to further provide insight of the proposed method. We first study the sensitivity parameters of the proposed vMFML parameters (Sect. \ref{sssection: Sensitivity analysis of the vMFML parameters}); Then, we conduct further experiments and analysis to discuss the influences of several CNN training related aspects, such as: (a) training datasets size (Sect. \ref{sssection: Impact of training dataset size}); (b) activation functions (Sect \ref{sssection: Influence of activation function}) and (c) normalization (Sect. \ref{sssection: Feature normalization}). Finally, we also analyze the limitations of the proposed method (Sect. \ref{sssection: Limitations of the proposed method}). In order to evaluate, we observe the accuracy and TAR@FAR=0.01 on the LFW benchmark. 
\subsubsection{Sensitivity analysis of the vMFML parameters}
\label{sssection: Sensitivity analysis of the vMFML parameters}
The  proposed vMFML features two parameters, namely the concentration $\kappa$ and the mean $\mu$, whose sensitivity is analyzed here.  In general, we can initialize $\kappa$ with a small value (\textit{e.g.}, 1) and learn it via backpropagation. However, we observe that the trained $\kappa$ may provide sub-optimal results, especially when it is trained with large number of CNN parameters and updated with the same learning rate. To overcome this, we set $\kappa$ to an approximated value as $\kappa = \sqrt{d/2}$ and set its learning rate by multiplying the CNN learning rate with a small value 0.001. An alternative choice is to set $\kappa$ to a fixed value for the entire training period, where the value will be determined empirically. 
Our best results are achieved with $\kappa=16$.
On the other hand, the parameter $\mu$ does not exhibit any particular sensitivity and learned in a similar way to other CNN parameters. 
\subsubsection{Impact of training dataset size}
\label{sssection: Impact of training dataset size}

\textit{Training dataset size} matters and plays important role to learn good feature representations with the deep CNN models  \cite{schroff2015facenet, zhou2015naive}. To investigate its effect on our FR method, first we train the \textit{Res-27 CNN model} (Sect. \ref{proposed_cnn_model}) with vMFML (Sect. \ref{ssec:vmfml}) using the well-known training datasets, such as the MSCeleb-1M \cite{mscelebguo16}, CASIA-Webface \cite{yi2014learning} and VGG Faces \cite{parkhi2015deep}. For the MSCeleb-1M \cite{mscelebguo16} dataset, we created different subsets by selecting certain minimum number of images per identity. Experiments with these datasets and subsets help us to understand the learning capacity of the proposed FR method and identify the training dataset requirements to achieve better performance. 
Table \ref{tab:analysis_discussion_db} presents the results, which shows that the proposed method learns reasonably well facial representations from a wide range of different sized datasets. Additional observations are as follows:
\begin{itemize}
%
\item performance does not improve significantly with the increase of number of images and identities. 
We achieved only 0.28\% improvement by enlarging the number of images ~3 times and number of identities ~5 times (compare the results between MSCeleb-100 \textit{min-samp/id} with MSCeleb-10 \textit{min-samp/id}). Perhaps, it is important to ensure larger number of samples per identity (\textit{e.g.}, 100).
%
\item synthesized images does not help significantly. By comparing CASIA \cite{yi2014learning} with pose-augmented-CASIA ($\approx$3 times larger dataset) \cite{masi16dowe}, we observe only 0.03\% accuracy improvement. Perhaps the synthesized image quality affects the results.
\item number of identities is equally important as the number of images. We observe that the VGG Face \cite{parkhi2015deep} dataset provides lower accuracy compared to the CASIA \cite{yi2014learning} dataset despite having $\approx$4 times more number of images (but $\approx$4 times less number of identities).
\end{itemize}
Therefore, the key finding from this experiment is that, while we can achieve reasonably good and comparable results with relatively smaller datasets (CASIA and MSCeleb-100 \textit{min-samp/id}), we can achieve better results by training with larger dataset which provides sufficient number of images per identity. 
%
\begin{table}[t]
\footnotesize
\centering
\caption{\footnotesize{Analysis of the influences from training databases with different sizes and numbers of classes. T@F denotes the \textit{True Accept Rate at a fixed False Accept Rate (TAR@FAR)}.}}
\label{tab:analysis_discussion_db}
\begin{tabular}{|c|c|c|c|c|}
\hline
\textit{\textbf{Min samp/id}} & \textbf{\textit{Size, Class}} &  \textbf{\begin{tabular}[c]{@{}c@{}}Acc\\\%\end{tabular}} & \textbf{\begin{tabular}[c]{@{}c@{}}T@F\\0.01\end{tabular}}\\ \hline
\hline \hline
MSCeleb-1M \cite{mscelebguo16} & & & \\
\hline
10 & 4.68M, 62.1K & 99.57 & 0.9963 \\ \hline
30 & 4.61M, 61.2K & 99.63 & 0.9971 \\ \hline
50 & 3.91M, 46.7K & 99.55 & 0.9960 \\ \hline
70 & 3.11M, 32.1K & 99.53 & 0.9956 \\ \hline
100 & 1.53M, 12.5K & 99.35 & 0.9928 \\ \hline
\hline \hline
CASIA \cite{yi2014learning} & 0.43M, 10.6K & 99.18 & 0.9902 \\ \hline
Pose-CASIA \cite{masi16dowe} & 1.26M, 10.6K & 99.21 & 0.9907 \\ \hline
VGG Faces \cite{parkhi2015deep} & 1.6M, 2.6K & 98.45 & 0.9787 \\ \hline
\end{tabular}
\end{table}
\subsubsection{Influence of activation function}
\label{sssection: Influence of activation function}
In order to observe the influence of \textit{activation functions}, we also used the ReLU activation instead of PReLU and observe that it decreases the accuracy by approximately 0.4\%. 
\subsubsection{Feature normalization}
\label{sssection: Feature normalization}
\textit{Feature normalization} \cite{batch_normalization, normfacewang2017, deepvisage_2017_ICCVW} plays a significant role in the performance of deep CNN models.
As discussed in Section \ref{int_discussion}, vMFML naturally integrates features normalization due to the SFR model (Sect. \ref{sssec:ffrm}) and the vMF distribution (Sect. \ref{vmf_distribution}). The features of our FR method are unit-normalized vectors, \textit{i.e.}, $\left \| \mathbf{x} \right \| =1$. We observed that when learned with the proposed vMFML these normalized features provides significantly better results than the un-normalized features learned with the Softmax loss, see Table \ref{tab:sensitivity_comp_loss} for the performance comparison of \textit{Softmax} vs \textit{vMFML}. 

The promising results achieved by the unit-normalized features (learned with vMFML) naturally raises the question - \textit{can the unit-normalized features improves the accuracy with any loss function?} 
To answer this question, we train the CNN with the unit-normalized features $\left \| \mathbf{x} \right \| =1$ and optimize the softmax loss under different settings in order to compare them w.r.t. vMFML. To explain the settings, we exploit the terms within the exponential of the nominators, \textit{i.e.}, $\mathbf{w}^T\mathbf{x} + b_{y_i}$ (Softmax) with $\kappa\mu^{T}\mathbf{x}$ (vMFML). Our observations are:
\begin{itemize}
\item $\mathbf{w}^T\mathbf{x} + b_{y_i}$: provides very poor results, because the CNN training fails to converge as it get stuck at arbitrary local minimas even if with a range of different learning rates (from 0.1 to 0.0001) and with/without applying the $L2$ regularization. Compare to the vMFML, the expectation from this setting is that the learned  $\mathbf{w}$ values are similar to the values observed from $\kappa\mu^{T}$, such that, $\left \| \mathbf{w} \right \| =\kappa$. That means, this setting verifies whether the learned weights can absorb $\kappa$ within it.
However, as the observed results suggest, it fails to do so. 
\item $\mathbf{w}^T(\mathbf{x}\;\kappa) + b_{y_i}$: CNN successfully trains and provides results which is closer to the vMFML.
For this modified Softmax $\kappa$ acts as a scalar multiplier, whereas for vMFML $\kappa$ signifies the shared concentration parameter value. Interestingly, this modified Softmax is the same as the \textit{L2-Softmax} loss recently proposed by \cite{l2ranjan2017}, which is motivated by the fact that the L2-norm of the features (learned with the Softmax loss) provides interesting information of the face image quality and attributes. Intuitively, the idea is to provide same attention to all face images regardless of their quality. Moreover, they interpret the multiplier (here $\kappa$) as a constraint on the features to lie on a hypersphere of a fixed radius. Note that, if bias is ignored then the difference among vMFML and L2-Softmax remains as $\mathbf{w}$ vs $\mu$. Besides, while the softmax loss imposes no constraints\footnote{Except L2 regularization which is an explicit settings for many ML problems} on the weights  $\mathbf{w}$, vMFML applies natural constraint on $\mu$, \textit{i.e.}, $\left \| \mathbf{\mu} \right \| =1$. 
\end{itemize}
Therefore, the key finding from the above observations is that the unit-normalized features are not sufficient alone and concentration parameter $\kappa$ of vMFML plays significant role to efficiently learn the CNN models.
%
%
%

\subsubsection{Limitations of the proposed method}
\label{sssection: Limitations of the proposed method}
\begin{figure}[]
\centering
\subfloat[LFW \cite{lfw_huang2014}, threshold distance 0.665]{\includegraphics[scale=0.30]{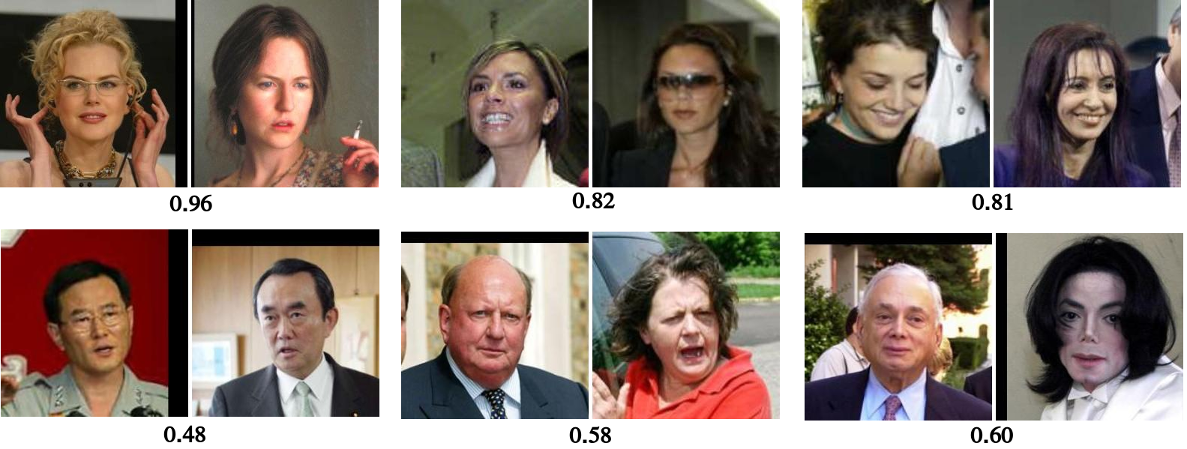}}
\hfill
\subfloat[CALFW \cite{calfwzheng2017}, threshold distance 0.66]{\includegraphics[scale=0.32]{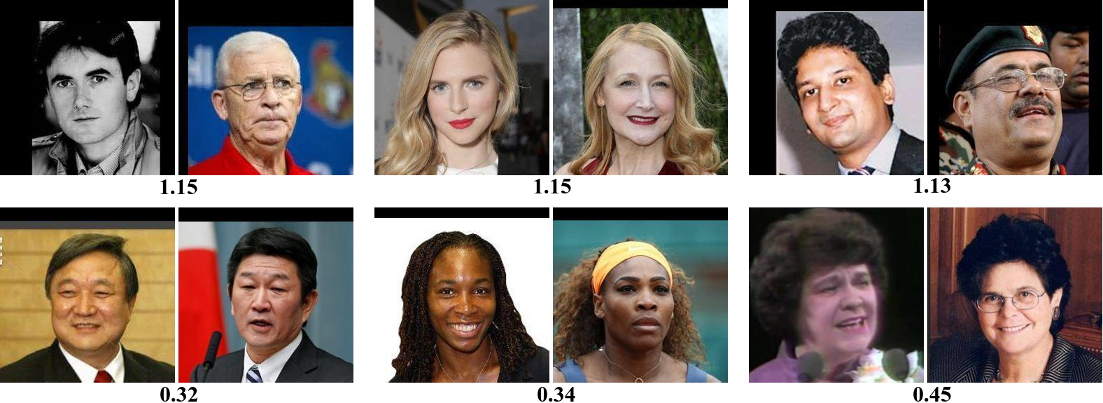}}
\hfill
\subfloat[YTF \cite{ytfwolf2011}, threshold distance 0.72]{\includegraphics[scale=0.305]{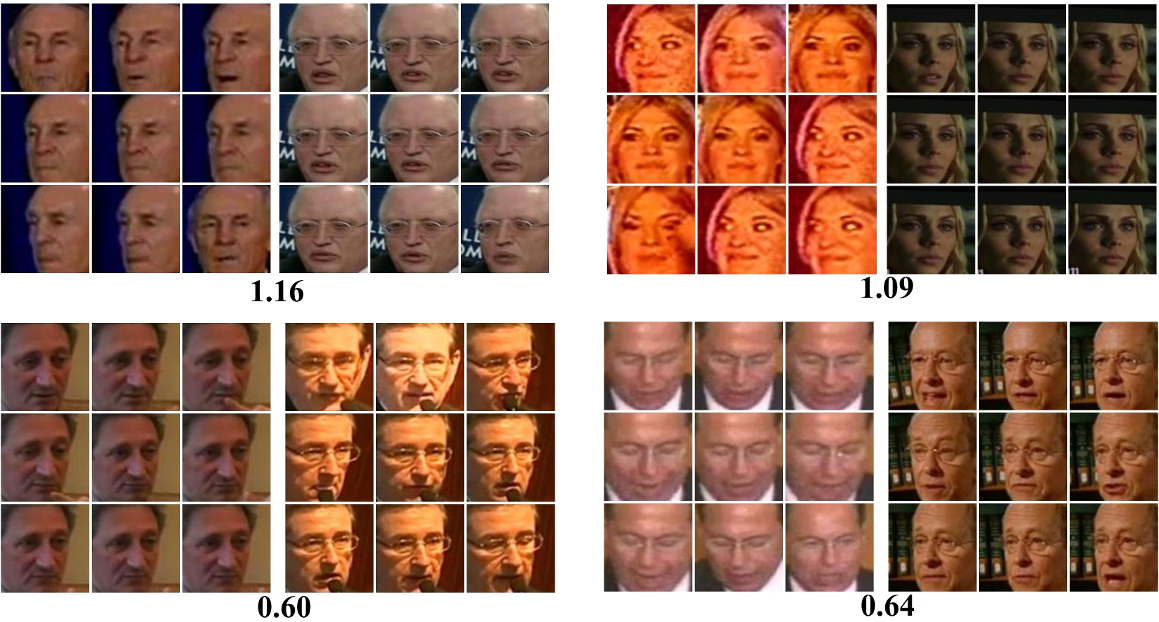}}
\hfill
\subfloat[CACD \cite{chen2015facecacd}, threshold distance 0.61]{\includegraphics[scale=0.32]{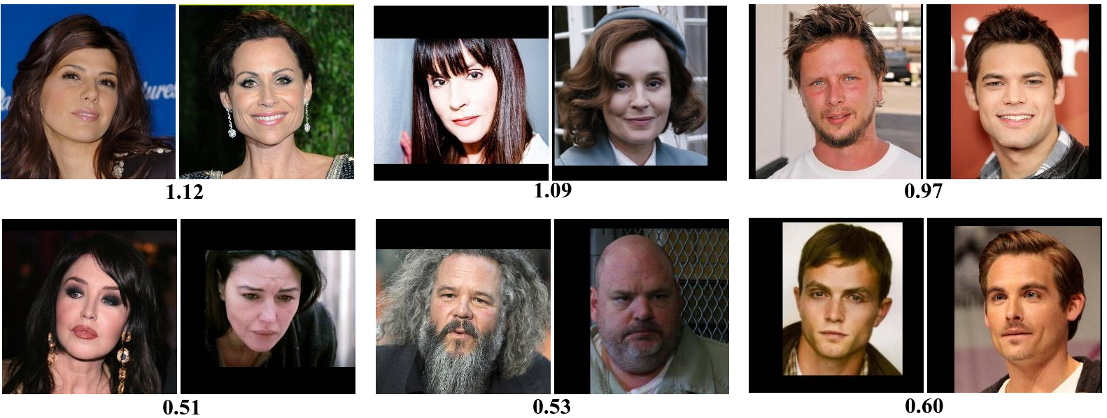}}
\hfill
\subfloat[IJB-A \cite{ijbaKlare2015}, threshold distance 0.49]{\includegraphics[scale=0.65]{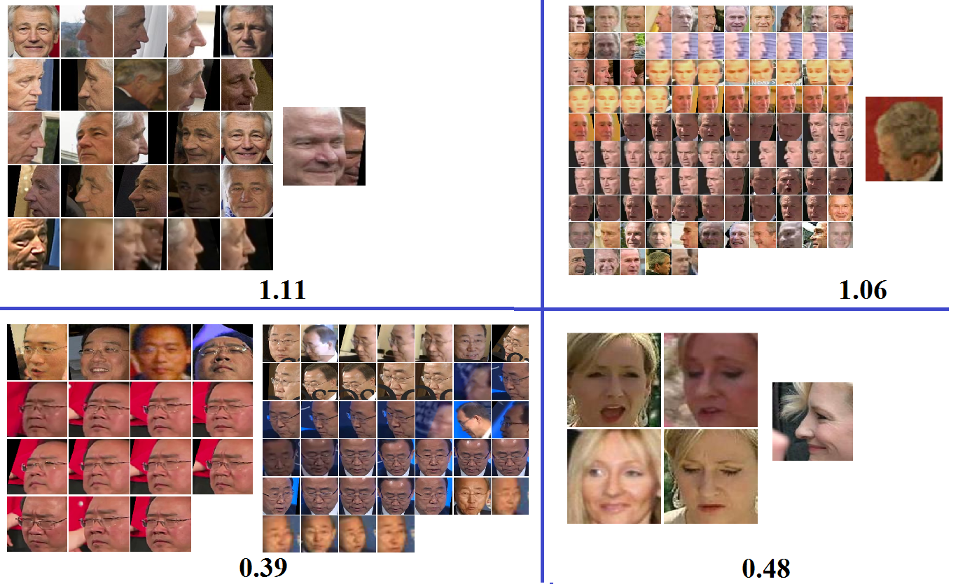}}
\footnotesize
\caption{Illustration of the errors made by our method on different datasets: (a) LFW \cite{lfw_huang2014}; (b) CALFW \cite{calfwzheng2017}; (c) YTF \cite{ytfwolf2011}; (d) CACD \cite{chen2015facecacd} and (e) IJB-A \cite{ijbaKlare2015}. 
The top row of each sub-figure provides examples of the false rejected pairs whereas the bottom row provides examples of the false accepted pairs.}
\label{fig:incorrect_dbs}
\end{figure}

Finally, we investigate the limitation of the proposed method on different datasets by observing the face image pairs for which the verification results are incorrect. Table \ref{tab:incorrect_res} provides the information about the number and type of incorrect cases, which indicate a higher ratio of false rejection compared to false acceptance. Figure \ref{fig:incorrect_dbs} provides illustrations of the top (selected based on the distance from threshold) examples of failure cases on different datasets. From an in-depth analysis of the erroneous results, our observations are as follows:
\begin{itemize}
\item On the LFW \cite{lfw_huang2014} failure cases, occlusion, variation of illumination and poor image resolution played important role. 
From the erroneous CALFW \cite{calfwzheng2017} pairs, we observe that poor image resolution appears as a common property.
Besides, the false rejected pairs are suffering from high age difference.
\item Failure cases on the YTF \cite{ytfwolf2011} dataset can be characterized by variations of illumination and poor image resolution. Besides, high pose variation plays an important role.
\item Most of the CACD incorrect results occurred due to falsely rejecting the similar face image pairs, which indicate that our method encounters difficulties to recognize the same person when the age difference is large. Besides, we observe that the variations of illumination commonly appear in the incorrect results.
\item The large number of errors on the IJB-A dataset can be characterized with high pose (mostly profile images, yaw angle more than 60 degrees) and poor image resolution. Unfortunately, both of these two reasons cause the failure of face and landmarks detection, which forced us to leave a large number of images without applying any pre-processing. Note that, our training dataset do not have any image for which the pre-processor failed to detect face and landmarks. Therefore, our method may perform poorly in case of the failure of the pre-processor.
\end{itemize}
From the above observations, we can particularly focus on several issues to handle in future, such as: (a) image resolution; (b) extreme pose variation (c) lighting normalization and (d) occlusions. 

\textit{Poor image resolution} is a common issue among the failure cases on different datasets. This observation is similar to the recent research by Best-Rowden and Jain \cite{faceimagequality2017}, which proposed a method to automatically predict the face image quality. They found that the results in the IJB-A \cite{ijbaKlare2015} dataset can be improved by removing low-quality faces before computing the similarity scores. Therefore, we can apply this approach for certain FR cases where multiple images are available for each identity. However, this approach will not work when only a single image is available per identity or when all available images have poor quality. In order to address this, we can incorporate face image super resolution based technique \cite{waveletsr2017} as a part of our pre-processor. The idea is to apply super-resolution to those images which have very low image quality score computed by techniques such as \cite{faceimagequality2017}.

\textit{Large facial pose} causes degradation of FR performance \cite{masi16dowe, huang2017beyond}. It has been addressed in recent researches \cite{abdalmageed2016face, masi2016pose, masi16dowe}, which proposed to overcome this problem by frontalization \cite{taigman2014deepface} and creating a larger training dataset with synthesized facial images of a range of facial poses \cite{masi16dowe, abdalmageed2016face}. However, the performance with above approaches depends on the quality of frontalization or synthesization.
Recently \cite{huang2017beyond} proposed a method to generate photo-realistic facial images. We can adopt this proposal within our pre-processor to overcome the large pose related issues. 

In order to deal with the \textit{occlusion} related problems we can adopt the deep feature interpolation based approach \cite{upchurch2016deep} to recover missing face attributes. Besides, to deal with eye-glass related occlusions, we can enlarge training images per identity by synthesizing faces with eye-glass using a generative method, such as \cite{larsen2015autoencoding}. 

In order to deal with the \textit{lighting} related issues, we can adopt recent techniques, such as \cite{li2017specular} for specular reflection problem and \cite{zhang2017improving} for shadow related problems.
The \textit{large age difference} in the same person image pair reduces the performance of our method. One possible way to overcome this problem is by augmenting (age based facial image synthesis with generative methods \cite{upchurch2016deep, antipov2017face}) more training images with the variation of ages. However, we must ensure that the synthesized images remains photo-realistic.

\begin{table}[]
\centering
\caption{Analysis of the number of incorrect results made by the proposed method on different FR datasets.}
\label{tab:incorrect_res}
\begin{tabular}{|c|c|c|c|}
\hline
\textbf{Dataset} & \textbf{\# of Errors} & \textbf{\begin{tabular}[c]{@{}c@{}}\# of False\\ Accept\end{tabular}} & \textbf{\begin{tabular}[c]{@{}c@{}}\# of False \\ Reject\end{tabular}} \\ \hline
LFW \cite{lfw_huang2014}             & 22 & 7 & 15 \\ \hline
CALFW \cite{calfwzheng2017}           & 368 & 111 & 257 \\ \hline
YTF \cite{ytfwolf2011}              & 197 & 13 & 184 \\ \hline
CACD \cite{chen2015facecacd}             & 29 & 3 & 26 \\ \hline
IJB-A \cite{ijbaKlare2015}            &  855 & 133 & 722 \\ \hline
\end{tabular}
\end{table}

\section{Conclusion}
\label{sec:conclusion}
We proposed a novel \textit{directional deep-features} learning method for FR by exploiting the concept of model-based clustering and combining it with the deep CNN models. 
%
First, we used the vMF mixture model as the theoretical basis to 
propose a statistical feature representation (SFR) model. Next, we developed an effective directional features learning method called vMF-FL, which formulated a novel loss function called vMFML. It has several interesting properties, such as: (a) learns discriminative features; (b) subsumes different loss functions and normalization techniques and (c) interprets relationships among parameters and object features. Extensive experiments on face verification confirms the efficiency and generalizability of vMF-FL as it achieved very competitive and state-of-the-art results on the benchmark FR datasets.
We foresee several future perspectives: (a) use the learned model to synthesize identity preserving faces and enhance training dataset and (b) explore SFR model with the generative adversarial network; and (c) apply it for other vision tasks (\textit{e.g.}, scene analysis), other domains (\textit{e.g.}, NLP, speech analysis) and other tasks (\textit{e.g.} clustering). Moreover, by ignoring the \textit{equal privilege assumption} one can further analyze the variations withing a class/cluster, which can be interesting for \textit{unsupervised} problems.
\appendices
\section{Experimental justification of the SFR model}
\label{secapp:proof_concept}
\begin{figure}[!t]
\centering
\subfloat[]{\includegraphics[scale=0.21]{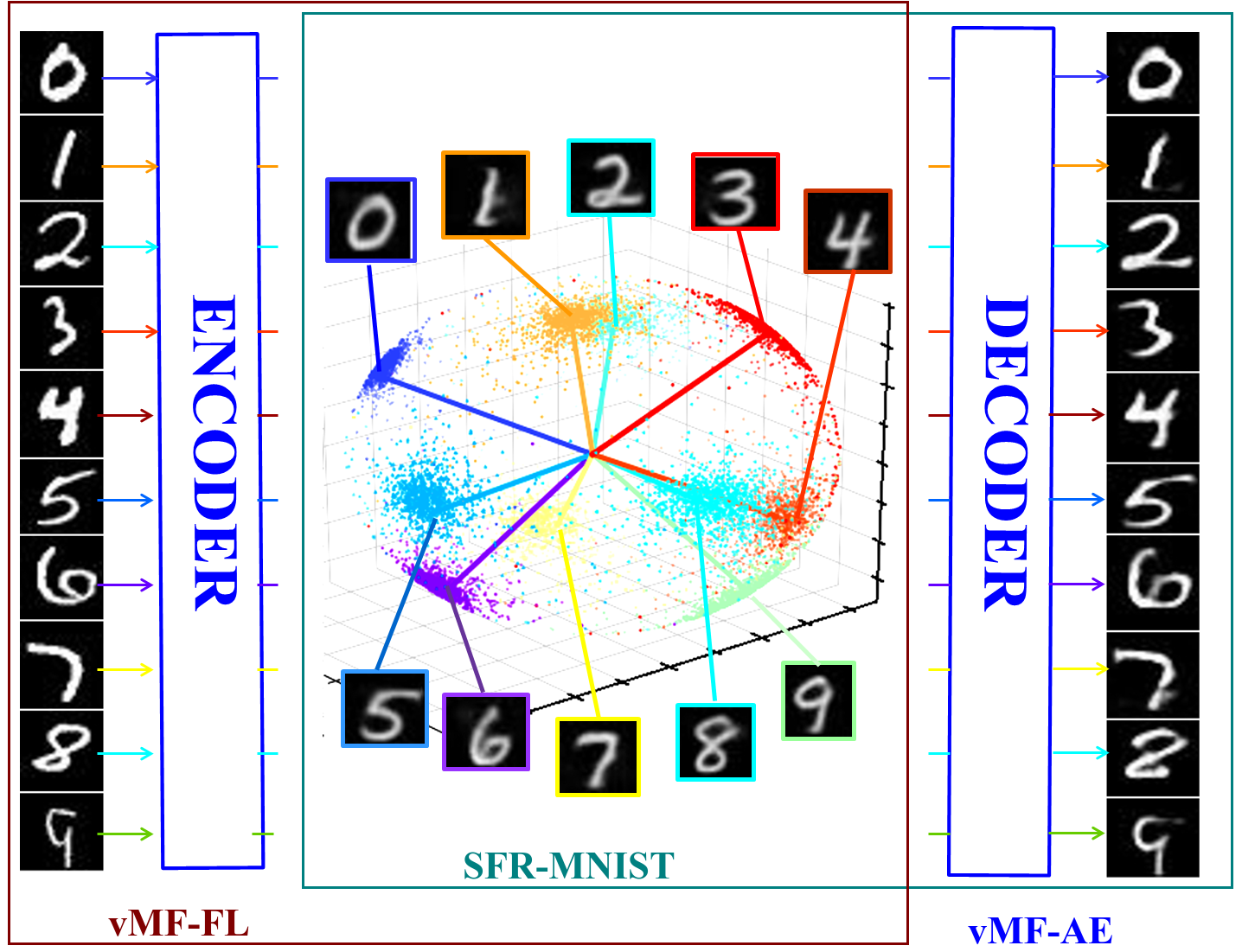}}
\hfill
\subfloat[]{\includegraphics[scale=0.38]{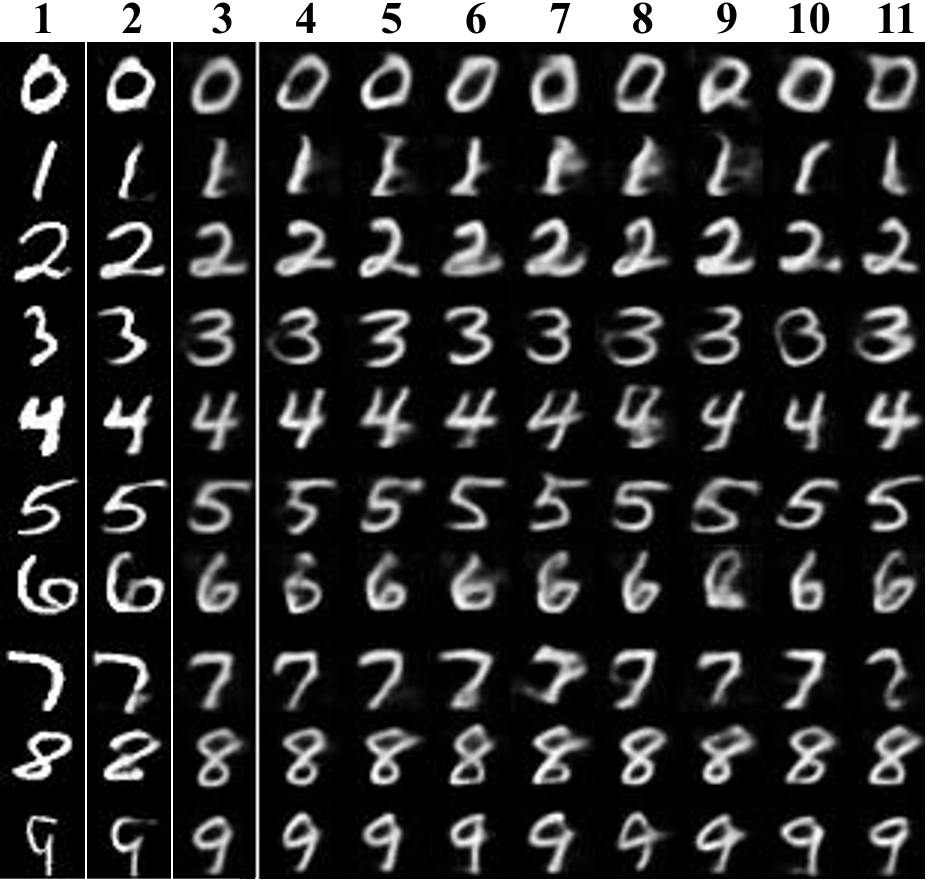}}
\caption{(a) illustration of the vMF-AE method with MNIST digits, where the vMF-FL (\textit{encoder/inverse-transformer}) learns discriminative 3D features and vMFMM for 10 digits-classes. In the 3D plot each dot represent features and each line represent the mean ($\mu_j$) of respective class $j$. The images around the 3D sphere illustrates the generated images from respective ($\mu_j$) and (b) illustration of the 2D images generated by the \textit{transformer/decoder} of the SFR model, where \textit{col-1} represents the original image, \textit{col-2} shows the image generated from the encoded features of the original image, \textit{col-3} shows the image generated from the vMF mean ($\mu_j$) of each class $j$, \textit{col-4-11} show 8 images from each digit-class $j$ generated from the vMF samples (sampled using the mean ($\mu_j$) and concentration ($\kappa_j$) of each class).}
\label{fig:proof_concept}
\end{figure}

In this experiment, we experimentally justify the proposed SFR model (Sect. \ref{sssec:ffrm}) and vMF-FL method (Sect. \ref{sssec:frlm}) by exploiting the concept of deep convolutional autoencoders \cite{autoencoderkingma2013}. To conduct this experiment with the MNIST digits \cite{lecun1998gradient}, we construct a deep autoencoder for this experiment, called vMF auto-encoder (vMF-AE). 
The \textit{encoder/inverse-transformer} of vMF-FL consists of a 7 layers (6 convolution layes, 1 fully connected layer) deep architecture and the \textit{decoder/transformer} of SFR model consists of a 4 layers (1 fully connected layer, 3 de-convolution layes) deep architecture.
Fig. \ref{fig:proof_concept} (a) illustrates the vMF-AE method and Fig. \ref{fig:proof_concept} (b) shows the generated samples from the SFR-MNIST model (column 4-11). 
%

vMF-AE combines the proposed SFR model with the vMF-FL method and learns in two steps. First it uses the vMF-FL method/block to simultaneously learn discriminative features (\textit{vMF feature}) representation and the vMF Mixture Model (\textit{vMFMM}) by optimizing the vMF mixture loss (Sect \ref{ssec:vmfml}). Next, it learns the \textit{decoder/transformer} of the SFR model to generate sample images from the features (learned by vMF-ML method) by optimizing pixel-wise binary-cross-entropy loss. When both learning tasks are done, we can generate 2D images of digits by sampling features from the learned vMFMM. While the column-3 of Fig. \ref{fig:proof_concept} (b) shows the images generated from different $\mu_j$ (\textit{learned vMFMM parameters}), column 4-11 of Fig. \ref{fig:proof_concept} (b) show several 2D images generated from the sampled vMFMM features. We believe that, these illustrations demonstrate the originality of the proposed SFR model and hence provides additional justification to the proposed vMF-FL method.

\section{Relationship with different loss functions and normalization method}
\label{secapp:rel_loss_functions}
%
%

\textit{\textbf{Center Loss \cite{centerlosswen2016}}} aims to enhance feature discrimination by minimizing the intra-class \textit{mean-squared} distances. 
It has the following form \cite{centerlosswen2016}:
\begin{equation}
\label{eq:center_loss}
\mathcal{L}_{Center} = \frac{1}{2} \sum_{i=1}^{N} \left \| \mathbf{f}_i - \mathbf{c}_{y_i} \right \|^2
\end{equation}
where, $\mathbf{f}_i$ and $y_i$ are the features and ground-truth class labels of the $i^{th}$ sample, $\mathbf{c}_{y_i}$ is the center of class $y_i$. By comparing Eq. \ref{eq:center_loss} and Eq. \ref{eq:vmfml_loss_soft}, we see that vMFML incorporates the \textit{cosine/angular} similarity with the term $\mu_{y_i}^{T}\mathbf{x}_i$, where $\mu_{y_i} = \frac{\mathbf{c}_{y_i}}{\left \| \mathbf{c}_{y_i} \right \|}$ and $\mathbf{x} = \frac{\mathbf{f}}{\left \| \mathbf{f} \right \|}$, and use it to compute the porterior probability followed by computing the loss. Note that, the higher the cosine similarity, the higher the probability and hence the lower the loss. In a different way, we can say that the loss in Eq. \ref{eq:vmfml_loss_soft} is minimized when the cosine similarity among the sample $\mathbf{x}_i$ and the mean $\mu_{y_i}$ of its true class $y_i$ is maximized. This indicates that, vMFML minimizes the intra-class distance by incorporating the distance computation task within its formulation.

The center loss is used as a supplementary loss to the softmax loss and the CNN learning task is achieved by joint (\textit{softmax+center}) optimization. On the other hand, the above comparison among vMFML and both \textit{softmax} loss and \textit{center} loss reveals that vMFML can take the advantage of both with a single loss function and save $M \times D$ parameters, where $D$ is the features dimension and $M$ is the number of classes in the training dataset. These additional parameters are used by the \textit{center} loss to externally learn and save the centers.

\textit{\textbf{Large-Margin Softmax Loss (LMSL) \cite{liu2016large}}} is defined as:
\begin{equation}
\footnotesize
\label{eq:lms_loss}
\mathcal{L}_{LMSL} = - \sum_{i=1}^{N} log\left ( \frac{e^{ \left \| \mathbf{w}_{y_i} \right \| \left \| \mathbf{f}_i  \right \| \psi \left ( \theta_{y_i} \right ) }}{e^{ \left \| \mathbf{w}_{y_i} \right \| \left \| \mathbf{f}_i  \right \| \psi \left ( \theta_{y_i} \right ) }
 + \sum_{j \neq y_i}  e^{ \left \| \mathbf{w}_j \right \| \left \| \mathbf{f}_i  \right \| cos \left ( \theta_j \right ) }} \right )
\end{equation}
where, $\psi \left ( \theta_{y_i} \right ) = (-1)^h cos(m \theta) - 2h$ with integer $h \in \left [ 0, m-1 \right ]$ and $\theta \in \left [ \frac{h \pi}{m}, \frac{(h+1)\pi}{m} \right ]$. $m$ denotes the margin, $\mathbf{f}_i$ is the $i^{th}$ image features, $y_i$ is the true class label, $\mathbf{w}_j$ is the weight corresponding to the $j^{th}$ class and $\theta_j$ is the angle between $\mathbf{w}_j$ and $\mathbf{f}_i$. 

vMFML has similarity to LMSL when $\left \| \mathbf{w}_{y_i} \right \| = \left \| \mu \right \| = 1, \; \forall\,y_i$ and $\left \| \mathbf{f}_i \right \| = \left \| \mathbf{x}_i \right \| = 1$. In this condition, LMSL requires $m\theta_{y_i} < \theta_j$ ($j \neq y_i$), \textit{i.e.}, the angle between the sample and its true class is smaller than the rest of the classes subject to a margin multiplier $m$. With vMFML, this can be achieved by using a higher $\kappa$ value (or multiply $\kappa$ with a scalar multiplier $m$) for the true class compared to the rest, \textit{i.e.}, $\kappa_{y_i} > \kappa_{j \neq y_i}$.
By considering the multiplier is equivalent to the margin $m$, we can rewrite Eq. \ref{eq:vmfml_loss_soft} as:

\begin{equation}
\label{eq:vmfml_loss_all_lmsl}
\mathcal{L}_{vMFML} = - \sum_{i=1}^{N} log \left ( 
\frac{e^{ m \kappa \mu_{y_i}^{T}\mathbf{x}_i }}{{e^{ m \kappa \mu_{y_i}^{T}\mathbf{x}_i } + \sum_{j \neq y_i}  e^{ \kappa \mu_{j}^{T}\mathbf{x}_i }}} \right )
\end{equation}
Note that, we do not use the Eq. \ref{eq:vmfml_loss_all_lmsl} in this work, because it does not strictly follow the statistical features representation model proposed in this paper. The A (angular)-softmax loss \cite{spherefaceliu2017} is a recent extension of the  LMSL, which replaces the notion of \textit{margin} with \textit{angular margin}, considers $\left \| \mathbf{w}_{y_i} \right \| = 1$ and provides an equivalent loss formulation as Eq. \ref{eq:lms_loss}. Therefore, Eq. \ref{eq:vmfml_loss_all_lmsl} provides the relationship among vMFML and A-softmax \cite{spherefaceliu2017} in a similar way as LMSL.
%
%

\textit{\textbf{Weight Normalization \cite{weight_normalization}}} proposed a reparameterization of the standard weight vector $\mathbf{w}$ as:
\begin{equation}
\label{eq:weight_norm}
\mathbf{w} = g \frac{\mathbf{v}}{\left \| \mathbf{v} \right \|}
\end{equation}
where, $\mathbf{v}$ is a vector, $g$ is a scalar and $\left \| \mathbf{v} \right \|$ is the norm of $\mathbf{v}$. This is related to the vMFML by considering the weight vector $\mathbf{w}$ as:
\begin{equation}
\label{eq:weight_norm_vmfml}
\mathbf{w} = \kappa \frac{\mu}{\left \| \mu \right \|} = \kappa \mu; \;\; \left [ \left \| \mu \right \| = 1 \right ]
\end{equation}
The comparison of Eq. \ref{eq:weight_norm} and Eq. \ref{eq:weight_norm_vmfml} reveals that vMFML incorporates the properties of weight normalization subject to normalizing the activations of the CNN layer which are considered as the features.
{
\bibliographystyle{ieee}
\bibliography{fr_dl}
}

\end{document}